\documentclass[runningheads]{llncs}

\PassOptionsToPackage{numbers, sort&compress}{natbib} 
\PassOptionsToPackage{numbers, sort&compress}{natbib}

\usepackage{eccv}

\usepackage{eccvabbrv}

\usepackage[utf8]{inputenc} %
\usepackage[T1]{fontenc}    %
\usepackage{url}            %
\usepackage{booktabs}       %
\usepackage{amsfonts}       %
\usepackage{nicefrac}       %
\usepackage{microtype}      %
\usepackage{graphicx}
\usepackage[export]{adjustbox} %
\usepackage{booktabs} %

\usepackage{amsmath}
\usepackage{amssymb}
\usepackage{mathtools}

\usepackage{enumitem}

\usepackage{stmaryrd}

\usepackage{tabularx}
\usepackage{makecell}

\usepackage{url}            %
\usepackage{booktabs}       %
\usepackage{amsfonts}       %
\usepackage{nicefrac}       %
\usepackage{microtype}      %
\usepackage{multirow}       %
\usepackage{algorithm}
\usepackage{algorithmic}

\usepackage{wrapfig}
\usepackage{comment}
\usepackage{amsmath,amssymb} %
\usepackage{symbols}
\usepackage{multirow}
\usepackage{pifont}%

\usepackage{amsmath,amsfonts,bm}

\def\1{\bm{1}}

\def\sp{space}

\def\rvc{{\mathbf{c}}}

\def\rvx{{\mathbf{x}}}

\def\vc{{\bm{c}}}

\def\vn{{\bm{n}}}

\def\mI{{\bm{I}}}

\DeclareMathAlphabet{\mathsfit}{\encodingdefault}{\sfdefault}{m}{sl}
\SetMathAlphabet{\mathsfit}{bold}{\encodingdefault}{\sfdefault}{bx}{n}
\newcommand{\tens}[1]{\bm{\mathsfit{#1}}}

\def\tM{{\tens{M}}}

\def\gL{{\mathcal{L}}}
\def\gM{{\mathcal{M}}}
\def\gN{{\mathcal{N}}}

\def\gT{{\mathcal{T}}}

\newcommand*{\ShowNotes}{} %
\definecolor{darkred}{rgb}{0.7,0.1,0.1}
\definecolor{darkgreen}{rgb}{0.1,0.7,0.1}
\definecolor{cyan}{rgb}{0.7,0.0,0.7}
\definecolor{dblue}{rgb}{0.2,0.2,0.8}
\definecolor{maroon}{rgb}{0.76,.13,.28}
\definecolor{burntorange}{rgb}{0.81,.33,0}
\definecolor{tealblue}{rgb}{0.212,0.459, 0.533}

\definecolor{mypink}{rgb}{0.93359375, 0.62109375, 0.83984375}

\definecolor{pp}{rgb}{0.43921569, 0.18823529, 0.62745098}
\definecolor{rr}{rgb}{0.5254902 , 0.00784314, 0.12941176}
\definecolor{bb}{rgb}{0.09019608, 0.23529412, 0.37647059}
\definecolor{yy}{rgb}{0.49803922, 0.3372549 , 0.0}
\definecolor{gg}{rgb}{0.02352941, 0.3372549 , 0.17647059}

\ifdefined\ShowNotes
  \newcommand{\colornote}[3]{{\color{#1}\bf{#2: #3}\normalfont}}
\else
  \newcommand{\colornote}[3]{}
\fi

\newcommand{\name}{{\textit{Tree-D Fusion}}}

\usepackage[multiple]{footmisc}

\newcolumntype{Y}{>{\centering\arraybackslash}X}

\definecolor{mybrown}{rgb}{0.87058824, 0.56078431, 0.01960784}
\definecolor{myblue}{rgb}{0.3372549 , 0.70588235, 0.91372549}
\definecolor{mypurple}{rgb}{0.8, 0.47058824, 0.7372549 }
\definecolor{myorange}{rgb}{0.835, 0.368, 0}
\definecolor{mygreen}{rgb}{0.00784314, 0.61960784, 0.45098039}
\definecolor{mygt}{rgb}{0.0078125 , 0.57421875, 0.40625}
\definecolor{mysp}{rgb}{0.84765625, 0.515625  , 0.0234375}
\definecolor{mycitecolor}{rgb}{0,0.08,0.45}
\definecolor{mygr}{rgb}{0.9607,0.9607,0.9607}
\definecolor{myoo}{rgb}{0.992,0.9176,0.9019}

\definecolor{myrr}{HTML}{AE031A}
\definecolor{mybb}{HTML}{0155B3}

\usepackage{pifont}
\usepackage[T1]{fontenc}
\usepackage[utf8]{inputenc}
\usepackage{pmboxdraw}
\usepackage{thm-restate}
\usepackage{natbib}
\usepackage[numbers,sort&compress]{natbib}

\usepackage[pagebackref,breaklinks,colorlinks,citecolor=eccvblue]{hyperref}

\usepackage{url}            %
\usepackage{booktabs}       %
\usepackage{amsfonts}       %
\usepackage{nicefrac}       %
\usepackage{microtype}      %
\usepackage{multirow}       %
\usepackage{algorithm}
\usepackage{algorithmic}
\usepackage{mathtools}%

\usepackage{arydshln}

\usepackage{symbols}
\usepackage{cite}
\usepackage{nicematrix}
\newcommand{\myparagraph}[1]{\vspace*{3pt}{\bf\noindent #1}}
\usepackage{orcidlink}
\usepackage{wrapfig}

\usepackage{amsthm}
\begin{document}
\title{ 
\name: Simulation-Ready Tree Dataset from Single Images with Diffusion Priors} %

\titlerunning{\name}

\author{Jae Joong Lee\inst{1}\orcidlink{0000-0002-0445-3141} \and
Bosheng Li\inst{1}\orcidlink{0009-0006-6490-1184} \and
Sara Beery\inst{2}\orcidlink{0000-0002-2544-1844} \and
Jonathan Huang\inst{3}\orcidlink{0009-0005-7490-8139} \and
Songlin Fei\inst{4}\orcidlink{0000-0003-2772-0166}\and
Raymond A. Yeh\inst{1}\orcidlink{0000-0003-4375-0680}\and
Bedrich Benes\inst{1}\orcidlink{0000-0002-5293-2112}
} 

\authorrunning{JJ Lee et al.}

\institute{Purdue University, Department of Computer Science \and
Massachusetts Institute of Technology, Department of Electrical Engineering and Computer Science \and
Google \and
Purdue University, Department of Forestry \& Natural Resources
\url{https://www.jaejoonglee.com/treedfusion/}
}

\maketitle
\begin{figure}
    \centering
    \includegraphics[width=0.9\linewidth]
    {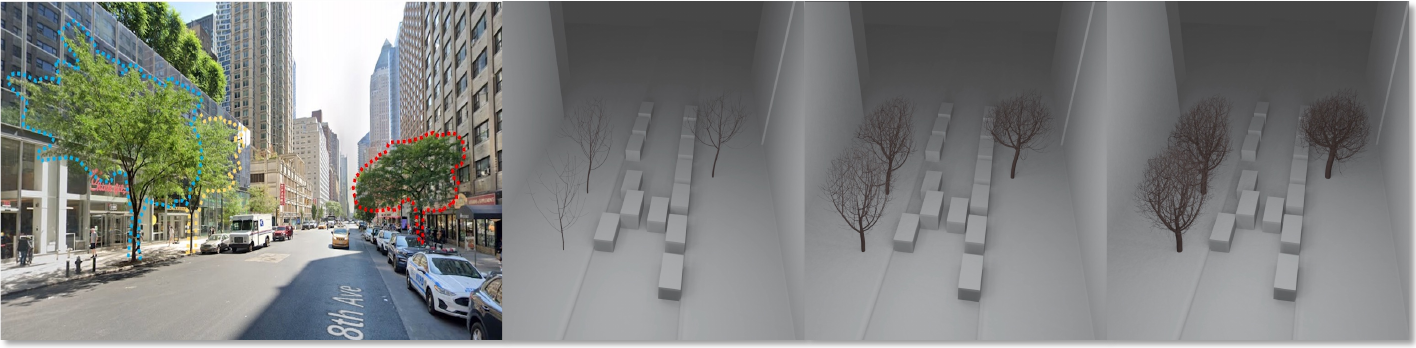}
    \caption{
    \name\ takes a single view image (left) and reconstructs a 3D simulation-ready tree model. The tree model can be used to simulate growth over time with a detailed branching structure with leaves. We provide a dataset of 3D reconstructed tree models from 600,000 Google Street View images.}
    \label{fig:teaser}
\end{figure}

\vspace{-0.9cm}
\begin{abstract}
We introduce \name, featuring the first collection of 600,000 environmentally aware, 3D simulation-ready tree models generated through Diffusion priors. Each reconstructed 3D tree model corresponds to an image from Google's Auto Arborist Dataset, comprising street view images and associated genus labels of trees across North America. Our method distills the scores of two tree-adapted diffusion models by utilizing text prompts to specify a tree genus, thus facilitating shape reconstruction. This process involves reconstructing a 3D tree envelope filled with point markers, which are subsequently utilized to estimate the tree's branching structure using the space colonization algorithm conditioned on a specified genus.
\end{abstract}

\section{Introduction}
Trees provide immense and essential value to human society and underpin diverse ecosystems worldwide~\cite{nowak18}. They cool the environment, improve air quality, capture carbon dioxide, produce oxygen, and have a positive effect on human physical and mental health~\cite{ziter2019scale,mcdonald2020value,yu2020preliminary,eisenman2019urban,blum2017contribution,bratman2019nature}. The complex effect of trees on the environment has been studied for centuries. Currently, computational models that seek to understand these relationships are hindered by a lack of data. In particular, acquiring phenotypical traits such as branching angles, wood volume, crown size, and trunk width is currently infeasible at a large scale. Simulation-ready 3D reconstructed trees (i.e., data that can be used in algorithms simulating tree development) that fit real-world data would enable the estimation of useful ecological quantities at scale. For example, the overall amount of wood or carbon sequestration, and would even allow for large-scale simulations of ``what-if'' scenarios, such as estimating shadow cover in both leaf-on and leaf-off scenarios at different times of the day or in different seasons~\cite{pan2022biomass,wang2019situ,zhang2022understory}.

Existing developmental models of vegetation typically rely on manual parameter tuning~\cite{Pwp:BOOK90,Palubicki:2009:STM,makowski2019synthetic,li2023rhizomorph}, or on a model parameter recovery that is fit to the data~\cite{stava2014inverse,Pirk:2012,niese2022procedural}. Digital datasets are scarce and usually available as unstructured point clouds~\cite{puliti2023forinstance,s19194188} or images~\cite{beery2022auto}. Providing realistically reconstructed trees that fit the input data at scale is yet to be achieved, as large-scale and geographically diverse data is needed to ensure that the synthetic versions accurately reflect the morphological diversity of real trees. Despite the abundance of tree images, converting them into 3D models has been difficult due to trees' complex shapes and self-occlusions from branches and leaves, which often hide significant portions of a tree. Thus, reconstruction methods often rely on biologically-based developmental algorithms that approximate missing parts~\cite{Li2021ToG,neubert2007approximate}.

We introduce~\name, the first tree dataset of its kind and scale,
containing 600,000 3D simulation-ready tree models grounded to individual real trees via street-level imagery and diffusion priors. The generation process is informed by genus-specific characteristics learned from the Auto Arborist Dataset (AAD)~\cite{beery2022auto} through a diffusion model trained specifically on tree images and captions. The result is a 3D envelope used by a genus-conditioned developmental model~\cite{li2023rhizomorph} that grows into this envelope while respecting the overall shape. The output is a reconstructed simulation-ready 3D tree model with a detailed branching structure and foliage.

We use the two Diffusion priors to reconstruct real trees from AAD, 
covering 23 cities across North America and over 300 genus categories. \name\ is validated against~\cite{qian2023magic123, liu2023zero1to3,tang2023dreamgaussian}. An example in~\cref{fig:teaser} left demonstrates how, from a single input image, our approach generates a model that can be used to simulate the growth of that specific real-world tree over time, even simulating the growth of leaves. Our dataset, \name\, contains 600,000 such models grounded on real-world trees, and we believe it will standardize benchmarking and advancements in forestry.

\section{Related Work}\label{sec:rw}
\myparagraph{Remote sensing for trees and forests.}
Tree and forest monitoring is an important topic in remote sensing, forestry, and ecology. Automated methods to measure canopy cover and forest height have been explored at large scales~\cite{crowther2015mapping,Loving2020, brandt2020unexpectedly, duncanson2018monitoring, aubry2021multisensor,lang2023high,locke2021residential}. Recent work seeks to do, modeling at the level of individual trees instead of forests, including individual tree detection, counting, and crown delineation~\cite{weinstein2019individual,stewart2021randcrowns,weinstein2021remote,plesoianu2020individual,ventura2022individual} and identifying the species of individual trees in both wilderness~\cite{harmon2023improving,marconi2022continental,graves2021data,fassnacht2016review,culman2021deep, safonova2019detection, hartling2019urban,plesoianu2020individual,weinstein2022capturing} and urban settings~\cite{ alonzo2014urban, sumbul2017fine,  zhang2012individual, hartling2019urban, aygunes2021weakly,beery2022auto,Wegner2016,Branson2018}. There is also increasing interest in understanding the structure of trees at an individual level to estimate tree attributes, including height, diameter at breast height (DBH), branching structure, and shadow estimation~\cite{jucker2022tallo}. In recent years, more sophisticated sensors such as airborne LiDAR have allowed for higher detail representations of individual tree shapes and structures, e.g.,~\cite{lines2022shape, zhou2020estimation, rosskopf2017modelling,bergseng2015assessing}. However, these sensor measurements are more expensive and cumbersome to collect at scale, especially at the level of individual trees. Thus, methods that allow for scalable measurement of tree geometry can significantly benefit these fields~\cite{tompalski2021estimating}.
Our dataset addresses this concern by offering a vast collection of realistic 3D tree models. 
Also, our data are grounded in real-world images and exhibit the most realistic reconstruction compared to recent methods, as demonstrated in~\cref{sec:quant_results}.

\myparagraph{Tree modeling} has been studied from a very different perspective in the computer graphics literature.
The generation of 3D tree geometric and behavioral models dates back to the seminal work of Lindenmayer, who introduced parallel string rewriting systems (L-systems) to simulate cellular subdivision~\cite{Lindenmayer:68}. L-systems were extended by Prusinkiewicz to capture branching patterns~\cite{Pwp:GI86} and later evolved~\cite{Pwp:BOOK90} into a full mathematical formalism that captures continuous development~\cite{Pwp:SIGG93,prusinkiewicz2018modeling}, plant signaling~\cite{Pwp:90a}, environmental interaction~\cite{Pwp:SIGG94}, and ecosystems~\cite{Deussen:98:SIGG}. Recent approaches focus on the simulation of interactions of tree models with wind~\cite{Pirk14ToG}, fire~\cite{hadrich2021fire}, climbing trees~\cite{hadrich2017interactive}, and ecosystems~\cite{niese2022procedural,makowski2019synthetic}. One of the most important open problems in tree modeling is reconstructing the developmental model from data~\cite{latentLsystem,Guo20ToG,Stava10CGF,mcquillan2018algorithms}. A developmental model that simulates an existing tree would allow for estimating its shape under different conditions and predicting its shape changes and the effect on the environment.

\myparagraph{Tree geometry reconstruction} methods can be classified according to the input data type, and the most prevailing are RGB images and point clouds. Tree reconstruction from point clouds involves creating a tree skeleton~\cite{Cardenas2022modeling} that captures tree topology and basic features (branching angles, branch length)~\cite{xu2007knowledge,livny2010automatic,Du2019adtree}. Other methods use voxels to retrieve the shape from particle flow~\cite{neubert2007approximate} or fitting geometric proxies to approximate the branch shape~\cite{hu2017efficient,hackenberg2015simpletree,aiteanu2014hybrid,Aitenau21CaG}.
Although tree geometry has been extensively investigated, only a limited number of studies have concentrated on foliage reconstruction~\cite{Bradley2013,Ando21PlantPhenomics,yin2016full,chen2021individual,chaurasia2017editable} or trees incorporating leaves~\cite{xie2018reconstruction,livny2011texture}.
RGB images are a much more widely accessible modality, so there is significant interest in reconstructing trees from RGB image input.
Some methods make use of incomplete information supplied via user input to identify the trunk and branches~\cite{Tan2007,tan2008single,quan2006image,cheng2007simple,liu2021single}. Other methods create 3D voxels with approximate information that control tree growth~\cite{Reche2004} or use multi-image input~\cite{guo2020realistic}. Recent deep learning approaches use transformer-based tree reconstruction to L-systems~\cite{latentLsystem}, single image-based tree reconstruction using convolutional networks~\cite{Li2021ToG}, statistical inference~\cite{Isokane2018CVPR}, or adversarial networks~\cite{liu2021single}. Our approach is inspired by an algorithm that generates a 3D geometric envelope completed by a developmental model~\cite{Li2021ToG}. Rather than a static proxy, we offer a detailed 3D envelope capturing the tree's shape more accurately.

\myparagraph{Single image 3D model reconstruction} has been an active area of research for a long time~\cite{dai2017shape,wang2018pixel2mesh,mahmud2020boundary,mescheder2019occupancy,chen2020bsp,peng2020convolutional,nash2020polygen,park2019deepsdf,Duggal_2022_CVPR}. Given RGB images, these methods aim to reconstruct common 3D model types such as mesh~\cite{wang2018pixel2mesh,xu2019disn, worchel2022multi},
voxels~\cite{choy20163d, yang2020robust, xie2020pix2vox++, wang2021multi, peng2022tmvnet}, or point clouds~\cite{fan2017point}.
Recently, diffusion models~\cite{sd1_5} for 3D generation applications~\cite{wang2023score,dreamfusion,wang2023prolificdreamer} have been used as priors for single image reconstruction~\cite{xu2023neurallift,deng2023nerdi,raj2023dreambooth3d,qian2023magic123,liu2023zero1to3,tang2023dreamgaussian}. 
These approaches significantly improve reconstruction quality for occluded parts in a given input image.

\begin{wrapfigure}[14]{r}{0.4\textwidth} 
\centering
    \begin{tabular}{ccc}
        \specialrule{.15em}{.05em}{.05em}
        Input Image & Front-view & Side-view\\
        \midrule
        \includegraphics[width=0.3\linewidth]{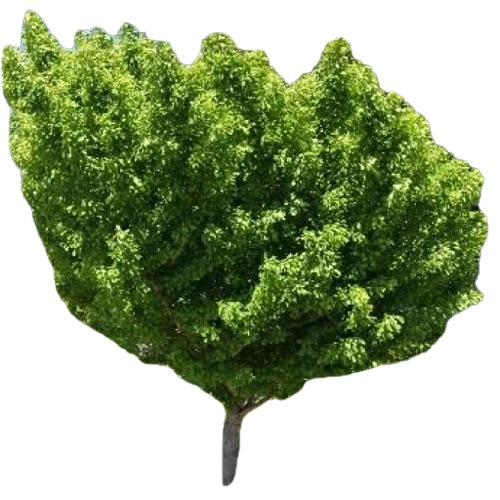}
         &
        \includegraphics[width=0.3\linewidth]{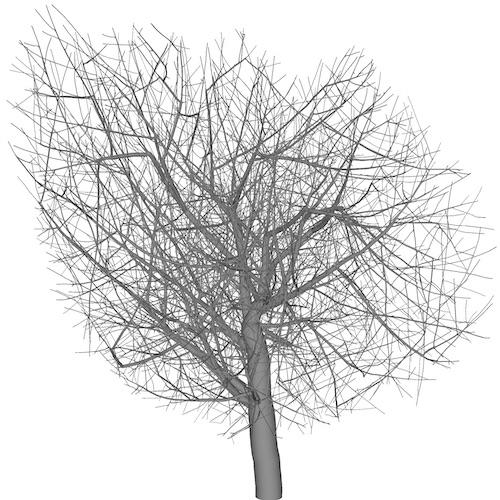}
        &
        \includegraphics[width=0.3\linewidth]{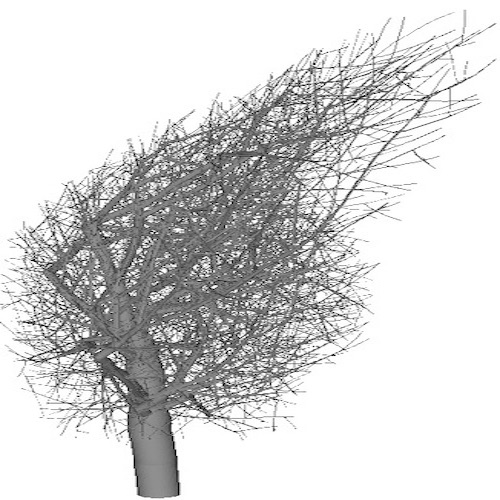}\\
        \includegraphics[width=0.3\linewidth]{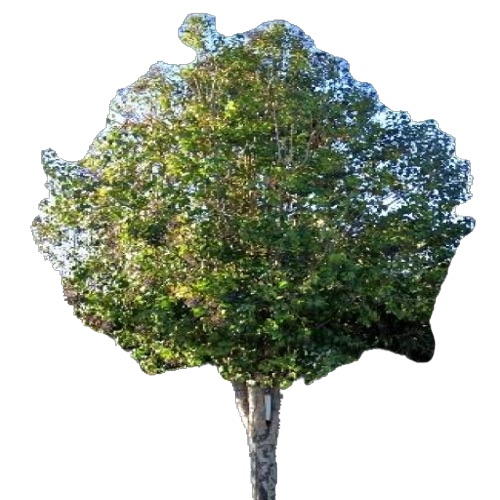}
         &
        \includegraphics[width=0.3\linewidth]{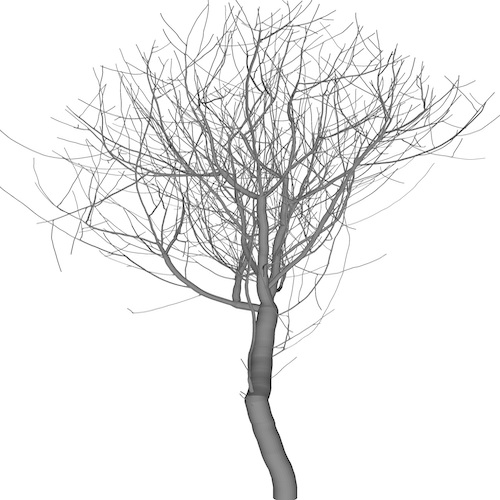}
        &
        \includegraphics[width=0.3\linewidth]{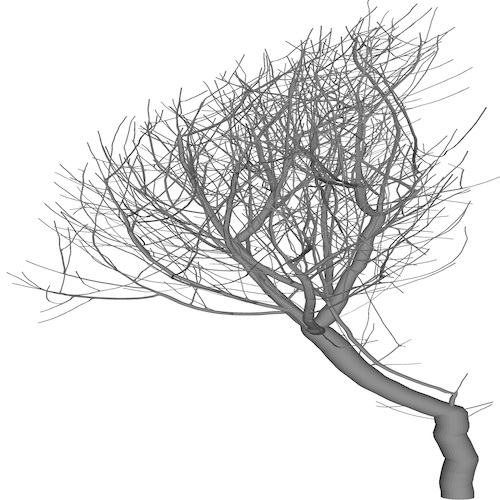}\\
        \specialrule{.15em}{.05em}{.05em}
    \end{tabular}
    \caption{
    Magic123~\cite{qian2023magic123} fails to capture trees' complex geometry.
    }
    \label{fig:app_compare}
\end{wrapfigure}

Despite these advancements, reconstructing trees from a single image remains challenging, and existing methods fail to create realistic 3D models of trees due to the intricate complexity of a tree's structure. 

\figref{fig:app_compare} demonstrates reconstructed trees using Magic123~\cite{qian2023magic123} that generates unrealistic tree geometry. These shortcomings of existing methods can be traced back to the inadequate representation of tree geometry in the diffusion models that are used as priors for the 3D shape, and that is the problem we aim to solve. \name\ addresses the challenge of predicting occluded branches from a single image by outlining the tree's contours and then using a species-specific procedural model~\cite{li2023rhizomorph} to grow the tree within these boundaries.

\section{Background}
We first briefly review diffusion models for 3D generation via score distillation~\cite{wang2023score}. As a probabilistic model, conditional diffusion aims to learn a model that can efficiently sample from the distribution $P(\mI| \rvc)$ of the image $\mI$ given some context~$\rvc$. This is done by learning a family of denoisers $D(\mI|\rvc;\sigma)$ at different Gaussian noise levels $\sigma$. As shown in the works~\cite{hyvarinen2005estimation,song2019generative},
the denoiser $D$ approximates the gradient field of the data log-likelihood, also known as the ``score function'':
\bea\label{eq:score}
\nabla_\mI \log P_\sigma(\mI|\rvc) \approx (D(\mI;\sigma)-\mI)/\sigma^2.
\eea

With this interpretation of a diffusion model,~\cite{wang2023score} proposed Score Jacobian Chaining (SJC) that applies the ``chain-rule'' to \textit{distill}  knowledge from a trained diffusion model. Specifically, 
SJC is applied to optimizing a NeRF~\cite{tensorf, dvgo} for generating 3D assets:
\bea
\frac{\partial P_\sigma(\mI|\rvc)}{\partial \theta} =  \underbracket{\frac{\partial P_\sigma(\mI|\rvc)}{\partial \mI}}_{\tt score} \frac{\partial \mI}{\partial \theta},
\eea
with image $\mI$ rendered from a NeRF with trainable parameters $\theta$.
However, since the rendered image is not noisy and the denoiser is trained on noisy images, na\"ively applying the score model leads to an out-of-distribution (OOD) problem.

Perturb-and-Average Scoring (PAAS)~\cite{wang2023score} is proposed to address this problem 
\bea\label{eq:paas}
\text{PAAS}(\rvx, \sigma) = \mathbb{E}_{\vn \sim \gN(0,\mI)} (D(\rvx + \sigma \vn;\sigma)-\rvx)/\sigma^2
\eea
to be used in place of the score. PAAS adds noise to the rendered image such that the data becomes in-distribution relative to the denoiser's training data.

We note that~\equref{eq:paas} is also known as score distillation in the concurrently proposed DreamFusion~\cite{dreamfusion} under a different mathematical formulation.
\name\ employs SJC/DreamFusion to distill the scores of diffusion models tailored for tree data, enabling the reconstruction of simulation-ready 3D tree models.

\begin{figure*}[t]
\includegraphics[width=\linewidth]{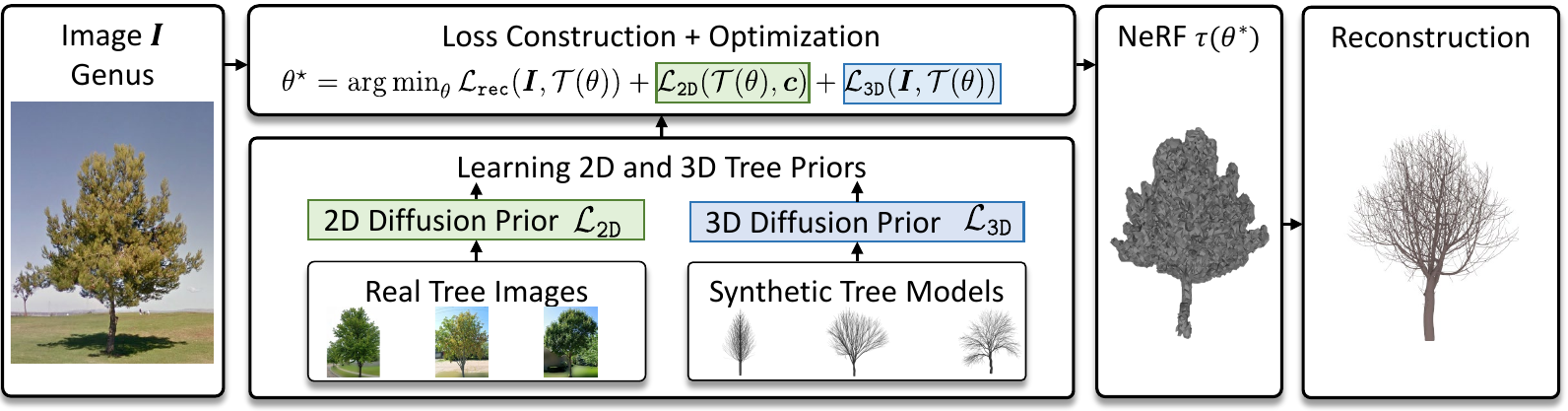}
\caption{The input to \name\ is an RGB image of a tree and its genus. 
To perform shape reconstruction, we minimize the 
loss function \wrt the NeRF parameter~$\theta$. The loss function is constructed from two diffusion models, StableDiffusion with Lora and Zero123, trained on real tree images and synthetic 3D tree models.
The output is an optimized NeRF $\tau(\theta^*)$, which is a detailed 3D tree envelope. We then populate the volume of $\tau(\theta^*)$ by markers based on the envelope and reconstruct trees by genus-conditioned space colonization algorithm.
}
\label{fig:pipeline}
\end{figure*}

\section{Simulation Ready Tree Dataset} 
We construct a large dataset of 600,000 reconstructed 3D trees that faithfully represent the 3D geometry and can simulate their growth, including environmental response to light and obstacles (see~\figref{fig:teaser}). 

The release of the AAD~\cite{beery2022auto}, featuring extensive tree images from Google Street View, aids the creation of 3D tree models from single images using a developmental model, capturing a specific instance of the tree's growth.
We leverage recent advances in diffusion models to reconstruct 3D trees. Given an input image, we first reconstruct a complete 3D volume of the tree using diffusion models as priors and fit the 3D volume by a developmental model conditioned on the tree genus. In particular, we use the space colonization model~\cite{Palubicki:2009:STM}, also used to reconstruct 3D tree geometry~\cite{li2023rhizomorph} for realistic tree growth.

An overview of our approach is illustrated in~\figref{fig:pipeline}. In the following, we describe how we train tree-specific diffusion models to be used as priors in~\secref{sec:tree_prior}, followed by the details of the 3D trees reconstruction in~\secref{sec:recon}. 

\subsection{Application}
\begin{figure}[t]
\centering
\includegraphics[height=3cm, width=0.9\linewidth]{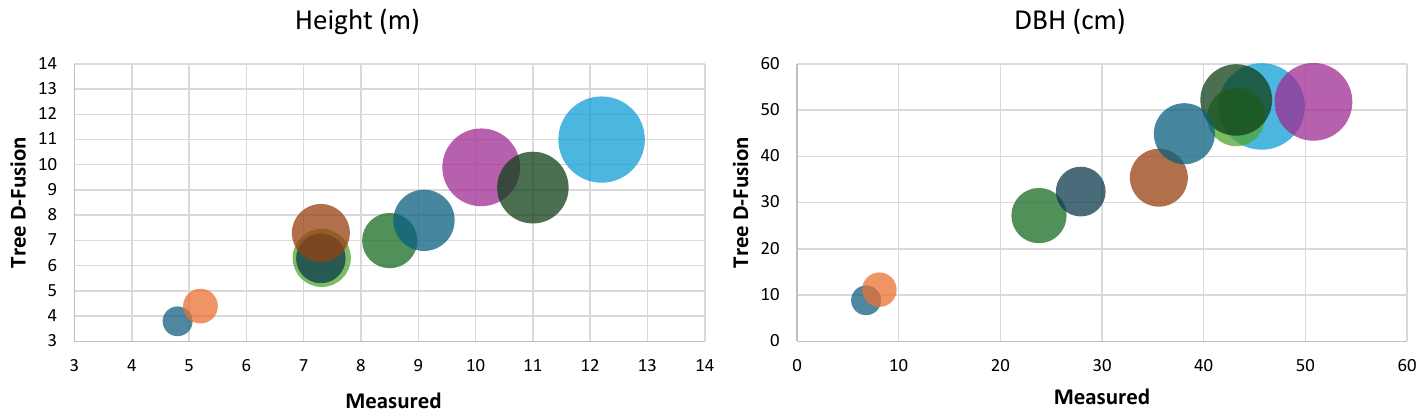}
\caption{Models of trees and the phenotypic characteristics derived from them: height and Diameter at Breast Height (DBH). The radius depicted illustrates the projected amount of shade. }
\label{fig:app_dbh_height}
\end{figure}

\begin{figure*}[hbt]
\centering
\includegraphics[height=1.5cm, width=0.9\linewidth]{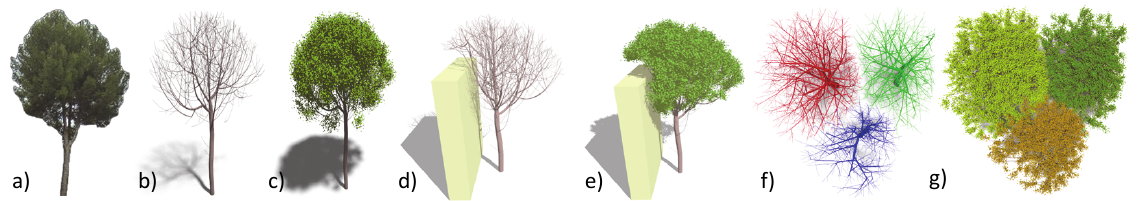}
\caption{The input image (a) is reconstructed into a digital twin (b-c) that responds to the environment such as the proximity to a wall (d-e). Growing three copies of the same tree shows their completion for space (f-g).}
\label{fig:env}
\end{figure*}

Our dataset, the first real-world 3D and 2D tree data at a large scale, seeks to standardize benchmarking and propel forestry deep learning, paralleling ImageNet's role in computer vision. It helps study distant trees, predicts tree diameter, and supports landscape design and forestry management. It also integrates into mapping and AR apps for enhanced 3D tree visualization.
Our dataset enables the creation of a 3D scene from a single Google Street View image, without the need for camera parameters, as demonstrated in~\cref{fig:teaser}. The image on the far left serves as the initial input, within which we delineate the three trees using dashed outlines. As time progresses, we model the development of these three trees. The final image depicts the mature trees, complete with foliage, ready for subsequent simulations like predicting urban heat island issues by estimating the total shadow area in the city.

\begin{wrapfigure}[9]{r}{0.25\textwidth} 
\centering
\includegraphics[height=1.2cm,width=\linewidth]{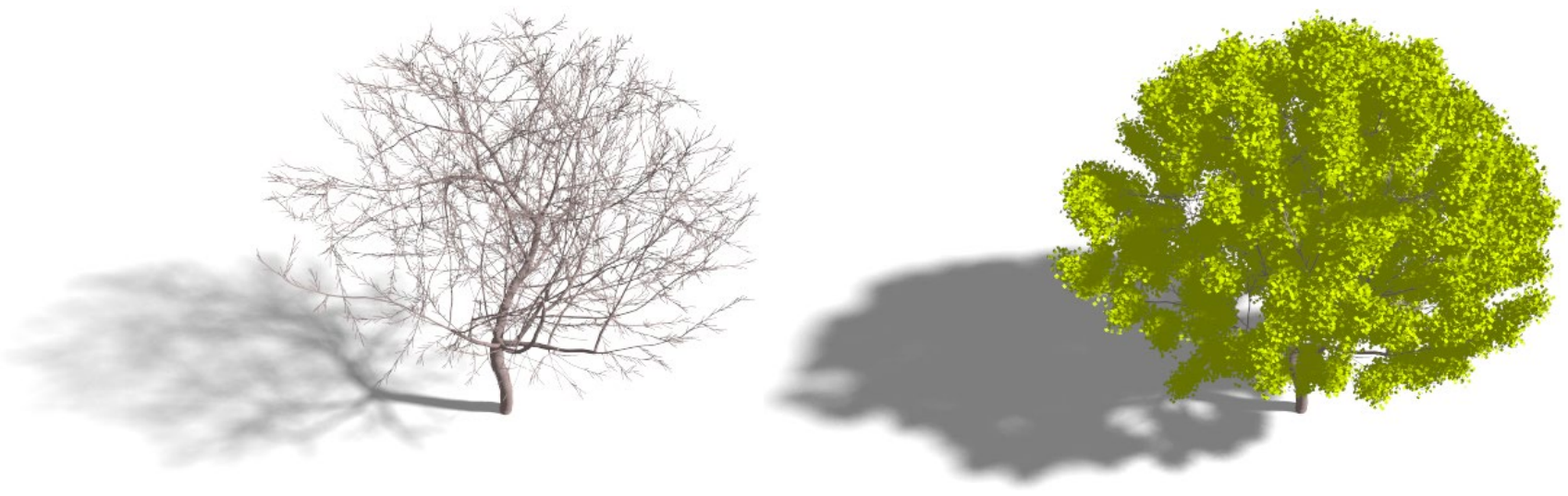}
\caption{Shadows cast by a tree without leaves (left) and with foliage (right).}
\label{fig:shadow_leaf_on_off}
\end{wrapfigure}

The reconstructed simulation-ready 3D trees are computational models that can be used in many practical applications. An example in \figref{fig:env}~(a) shows a tree from Google Street View images and its reconstructions (b-c). We then put a virtual object and simulate the tree's growth close to a wall (d-e), and we copy it twice and show its competition for space (f-g).

We quantitatively evaluate other phenotypical traits, such as tree height, the shadow area (as pixel-to-meter ratio), and the Diameter at Breast Height (DBH) commonly used in the forest industry (see~\figref{fig:app_dbh_height}). 
On average, our estimated DBH is predicted to be 16.4\% (4.0 cm) larger, with a standard deviation of 11.6\% (2.7 cm). Additionally, our model predicts a height of 12.5\% (1.0 m) shorter than the actual, with a standard deviation of 6.7\% (0.6 m). 
An example in~\figref{fig:shadow_leaf_on_off} shows the effect of shadow cast with and without leaves.

\begin{figure}[t]
\centering
\includegraphics[height=1.5cm, width=\linewidth]{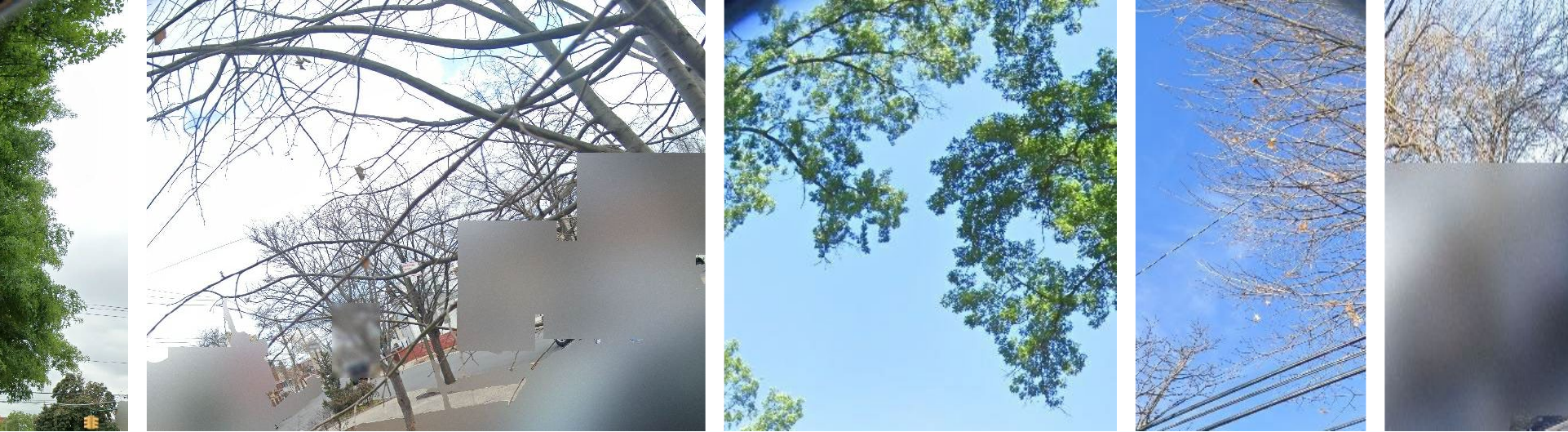}
\caption{Low-quality images from the AAD are excluded in~\name.}
\label{fig:app_noisy}
\end{figure}

\subsection{Training Tree-Specific 2D and 3D Diffusion Prior}\label{sec:tree_prior}
\myparagraph{Tree specific 2D prior.} In order to obtain an effective 2D prior, we need a diffusion model $P_{\tt 2D}(\mI|\vc;\sigma)$ that is trained to generate tree images $\mI$ given a corresponding text caption $\vc$. We utilize the AAD~\cite{beery2022auto}, which includes around one million real tree images. 

When training with these noisy labels, we observe that the diffusion model generalizes poorly and does not generate high-quality tree images. The dataset also contains images with heavy light exposure, extensive blurry images resulting from privacy filters, and low-image quality of distant trees (see~\figref{fig:app_noisy}). We semi-manually removed noisy images from the AAD to train this diffusion model. First, we divide the image into patches and measure the variance of the Laplacian in each patch. This approach measures image sharpness, averaging these values to exclude very blurry images for manual review. We will release this data along with this manuscript. 

We train a tree-genus-conditioned diffusion model using Low-Rank Adaptation (LoRA)~\cite{hu2021lora} on top of the pre-trained Stable Diffusion 1.5 version~\cite{sd1_5} as LoRA~\cite{hu2021lora} is more data efficient. We create image captions using BLIP~\cite{li2022blip} for genus conditioning and prefix the word ``tree'' within the caption with the specific genus annotation from AAD. For example, BLIP outputs the caption of {\tt ``A tree next to a house.''}. The caption will be {\tt A \{genus\} tree next to a house}, with \{genus\} replaced by the name of the genus. 

\myparagraph{Tree specific 3D prior.} To train the 3D prior, $P_{\tt 3D}(\mI_{\pi}| \mI, \pi;\sigma)$, we need data that contains paired images $(\mI, \mI_{\pi})$ of the same tree at two different viewing angles, and their corresponding viewing angle $\pi$. 

\begin{figure}[t]
\centering
\includegraphics[width=0.8\linewidth]{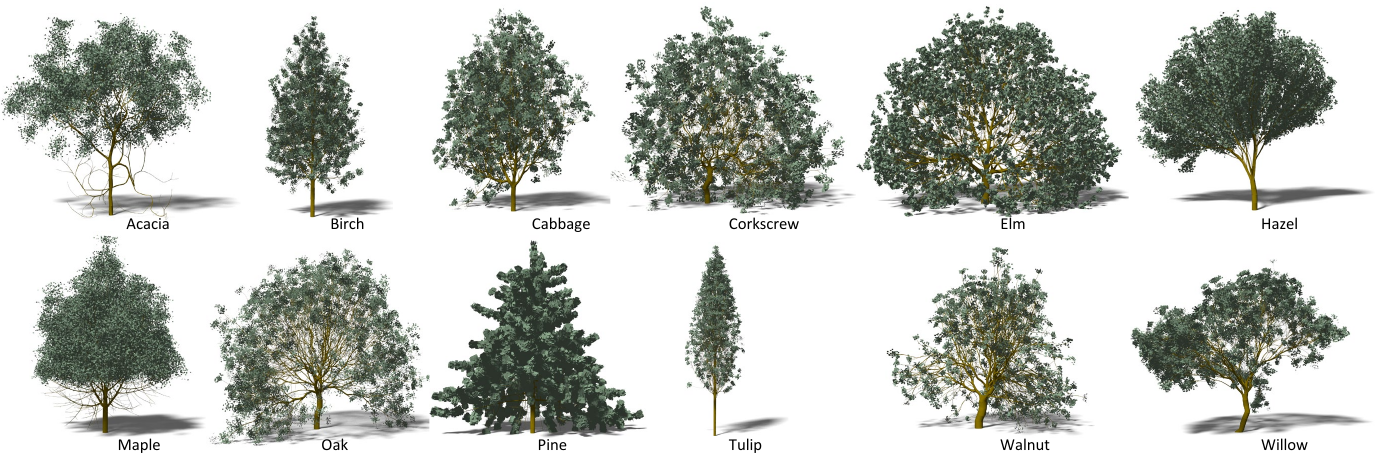}
\caption{Examples of synthetic trees used for training the 3D prior.}
\label{fig:app_syn}
\end{figure}

To obtain such data, we generated synthetic 3D tree models following~\citet{latentLsystem} over 12 different types of trees: Pine, Maple, Oak, Acacia, Birch, Cabbage, Corkscrew, Elm, Hazel, Tulip, Walnut, and Willow (some shown in~\figref{fig:app_syn}).

We produce 2,000 synthetic trees for each of the 12 tree categories. We train the diffusion model of Zero123~\cite{liu2023zero1to3} from scratch for 1,000 iterations. We train this model to be genera-agnostic, i.e., not conditioned on the genera, because our synthetic trees do not cover all the genera in the AAD dataset. For example, there are more than 100 genera of trees in San Jose. For training this 3D prior, we use tree geometry information, which includes 12 rendered views, each 30 degrees apart, per tree 3D object. 

\subsection{Single Image Tree Shape Reconstruction}\label{sec:recon}
We propose a specialized approach for reconstructing trees based on the generic 3D shape reconstruction method of Magic123~\cite{qian2023magic123}. We use our trained tree-specific diffusion models (\secref{sec:tree_prior}) and eliminate the depth and normal smoothness loss terms from Magic123 that we found not applicable to tree reconstruction.

\myparagraph{Problem formulation.} Given an tree image $\mI$ with a text prompt $\vc$, we formulate shape reconstruction as an optimization problem:
\bea\label{eq:all}
\min_\theta \gL_{\tt rec}(\mI, \gT(\theta)) + \alpha \gL_{\tt 2D}(\gT(\theta), \vc) + \beta\gL_{\tt 3D}(\mI,\gT(\theta)),
\eea
where $\gT(\theta)$ denotes the underlying 3D model, Instant-NGP~\cite{muller2022instant} in our case, with parameters $\theta$, $\gL_{\tt rec}$ denotes a reconstruct loss, $\gL_{\tt 2D}$ denotes a loss based on the 2D diffusion prior, $\gL_{\tt 3D}$ denotes a loss based on the 3D diffusion prior, and $\alpha$ and $\beta$ are weights for each of the priors.
 
\myparagraph{Loss definitions.} \textit{Reconstruction loss $\gL_{\tt rec}$:} As in Magic123, $\gL_{\tt rec}$ encourages the rendered image $\gT_\pi(\theta)$ from the view $\pi$ to match the input image $\mI$. The loss consists of two terms:
\bea%
\gL_{\tt rec} = \lambda_{\tt rgb} \norm{\gM(\mI)\odot (\mI-\gT_\pi(\theta))}_2^2
+ \lambda_{\tt mask}\norm{\gM(I)-\tM(\gT_\pi(\theta))}_2^2.
\eea
The first term calculates the difference between the pixels of the image within the segmentation mask of the tree. The symbol $\odot$ stands for element-wise multiplication. 
The second term encourages the 3D occupancy to be consistent with the 2D tree mask obtained from the segmentation model. To do this, the algorithm extracts a mask $\tM$ that indicates the occupancy at each pixel by integrating the volume density from the 3D model $\gT(\theta)$. We use the 
Segment Anything Model (SAM) for 2D instance segmentation masks~\cite{kirillov2023segany}.

\textit{2D diffusion prior loss $\gL_{\tt 2D}$:} A pre-trained diffusion model provides a prior for the non-observed parts of the tree. Specifically, we minimize the expected negative log probability of the diffusion model from different viewpoints $\pi'$, \ie,
\bea
\gL_{\tt 2D}(\gT(\theta), \vc) = \mathbb{E}_{\pi'} \left[-\log P_{\tt 2D}(\gT_{\pi'}(\theta)|\vc;\sigma)\right].
\eea
Here, $P_{\tt 2D}$ corresponds to the adapted diffusion model that specializes in tree image generation (\secref{sec:tree_prior}).

\textit{3D diffusion prior loss $\gL_{\tt 3D}$} utilizes another pre-trained diffusion model to provide a prior for tree generation. In contrast to our 2D prior, the 3D prior's diffusion model is trained to be 3D aware. Instead of conditioning on a text prompt, this diffusion model is trained to generate the image at a particular viewpoint $\pi'$ of a reference image $\mI$. Once again, we minimize the expected negative log probability of the diffusion model $P_{\tt 3D}$ over different viewpoints $\pi'$, \ie,
\bea
\gL_{\tt 3D}(\mI,\gT(\theta)) = \mathbb{E}_{\pi'} \left[-\log P_{\tt 3D}(\gT_{\pi'}(\theta)| \mI, \pi';\sigma)\right].
\eea

With the loss functions defined, we then
solve the optimization problem in~\equref{eq:all} using Adam~\cite{kingma2015adam}, using 
PAAS (Eq.~\ref{eq:paas}) to 
compute the gradient of $\gL_{\tt 2D}$ and $\gL_{\tt 3D}$ \wrt to $\theta$. %
We use a learning rate of 0.001 and optimize for 2,000 iterations without weight decay. We set $\lambda_{\tt rgb}=5.0$, $\lambda_{\tt mask}=20.0$, $\alpha=1.0$ and $\beta=8.0$.

\subsection{3D Simulation-Ready Tree Reconstruction}
\begin{wrapfigure}{r}{0.3\textwidth} 
\centering
\includegraphics[width=\linewidth]{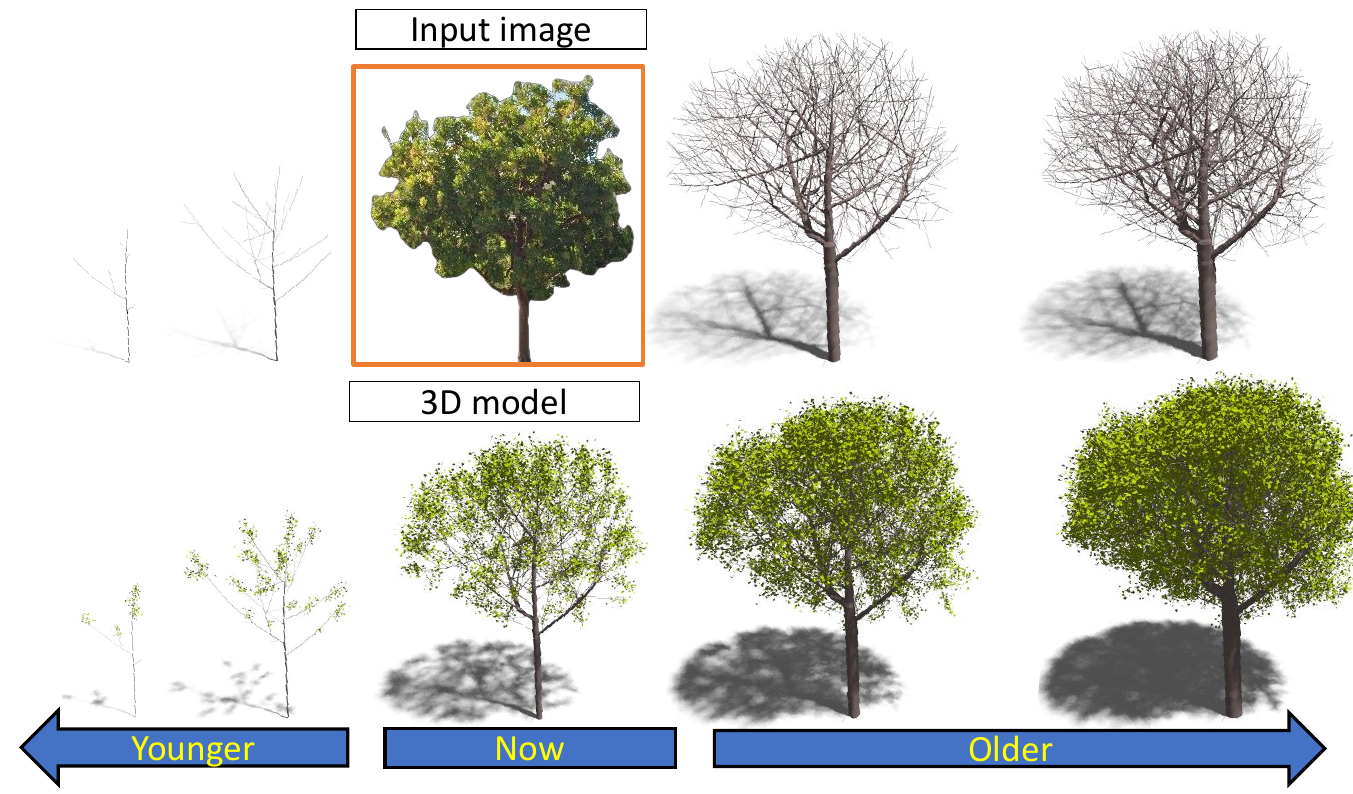}
\caption{The developmental model captures the real-tree shape (red frame) as a single instance in time. It allows us to look back and forward in time. }
\label{fig:growth}
\end{wrapfigure}

After computing the tree’s 3D geometry, we extract its 3D volume and use existing methods to create a simulation-ready 3D tree.
This algorithm is parameterized to generate the tree model. While various developmental models of trees (e.g., \cite{stava2014inverse,Pwp:BOOK90}) could be used, these are not suited for fitting the tree into a particular shape. Instead, we use the realistic tree growth algorithm~\cite{li2023rhizomorph} that extends the previous space colonization algorithm~\cite{Palubicki:2009:STM} for fitting the tree shape.
The algorithm~\cite{li2023rhizomorph} is conditioned on tree genus, which prescribes branching angles, internode lengths, branch radii, and other parameters~\cite{de1988plant,Weber95Sigg}. Environmental sensitivity is simulated by following existing works~\cite{Pwp:SIGG94,Pirk:2012}. ~\figref{fig:env} shows a reconstructed tree under varying conditions and~\figref{fig:growth} shows the %
simulation of the tree over time.

With the reconstructed 3D tree models from the Auto Arboist Datset~\cite{beery2022auto}, we aim to share the first real-world large 3D and 2D tree dataset. 
Our dataset can standardize benchmarking in forestry deep learning, similar to ImageNet’s impact on computer vision. This will advance research, aid in studying inaccessible trees, manage urban heat islands, and support landscape design and forestry management. Additionally, it can enhance map and augmented reality applications by displaying 3D trees.

\begin{table}[t]
    \caption{ICTree is the perceived realism scores of generated trees. \name\ shows an average improvement of 44.83\%$\pm$25.9\%.}   
    \label{tab:ictree_table}
    \setlength{\tabcolsep}{4pt}
    \centering
    \scriptsize
    \resizebox{\linewidth}{!}{%
    \begin{tabular}{cccccc}
    \specialrule{.15em}{.05em}{.05em}
    ICTree~\cite{ICTree}$\uparrow$ & DGauss~\cite{tang2023dreamgaussian} &  Magic123~\cite{qian2023magic123} & Zero123~\cite{liu2023zero1to3} & RBV~\cite{Li2021ToG} & Ours\\ %
    \midrule
    Cupressus & 0.46$\pm$0.03 & 0.45$\pm$0.05 & 0.46$\pm$0.04 & 0.75 $\pm$ 0.09 & \bf{0.65} $\pm$ 0.07 \\
    Magnolia & 0.46$\pm$0.04 & 0.45$\pm$0.03 & 0.46$\pm$0.03 & 0.616 $\pm$ 0.16 & \bf{0.71} $\pm$ 0.08\\
    Pinus &  0.44$\pm$0.04 & 0.46$\pm$0.03 & 0.44$\pm$0.04 & 0.627 $\pm$ 0.16 & \bf{0.69} $\pm$ 0.08\\
    Ligustrum & 0.45$\pm$0.04 & 0.45$\pm$0.03 & 0.45$\pm$0.04 & 0.665 $\pm$ 0.16 & \bf{0.71} $\pm$ 0.07\\
    Cinnamomum & 0.45$\pm$0.02 & 0.47$\pm$0.02 & 0.46$\pm$0.02 & 0.658 $\pm$ 0.16 & \bf{0.71} $\pm$ 0.08 \\
    \bottomrule
    Total  & 0.45$\pm$0.03 & 0.45$\pm$0.03 & 0.45$\pm$0.03 & 0.67 $\pm$ 0.12 & \bf{0.71} $\pm$ 0.05\\
    \specialrule{.15em}{.05em}{.05em}
    \end{tabular}
    }
\end{table}

\begin{table}[t]
    \caption{LPIPS~\cite{lpips} between a frontal view of a tree envelope from RBV~\cite{Li2021ToG} and \name\ shows 20.21\%$\pm$8.89\% improvement.}
    \label{tab:lpips_table}
    \setlength{\tabcolsep}{4pt}
    \centering
    \scriptsize
    \resizebox{\linewidth}{!}{%
    \begin{tabular}{cccccc}
    \specialrule{.15em}{.05em}{.05em}
    LPIPS~\cite{lpips}$\downarrow$ &  DGauss~\cite{tang2023dreamgaussian} &  Magic123~\cite{qian2023magic123} & Zero123~\cite{liu2023zero1to3} & RBV~\cite{Li2021ToG} & Ours\\
    \midrule
    Cupressus  &  0.68$\pm$ 0.02 & 0.68$\pm$ 0.02 & 0.69$\pm$0.02 & 0.53 $\pm$ 0.01 & \bf{0.52} $\pm$ 0.02\\   
    Magnolia   & 0.73$\pm$ 0.03 & 0.72$\pm$ 0.03 & 0.73$\pm$0.03 & 0.59 $\pm$ 0.05 & \bf{0.55} $\pm$ 0.06\\
    Pinus      & 0.70$\pm$ 0.01 & 0.69$\pm$ 0.02 & 0.70$\pm$0.02 & 0.55 $\pm$ 0.04 & \bf{0.50} $\pm$ 0.05\\
    Ligustrum  & 0.74$\pm$ 0.03 & 0.74$\pm$ 0.02 & 0.74$\pm$0.03 & 0.59 $\pm$ 0.04 & \bf{0.56} $\pm$ 0.05\\
    Cinnamomum & 0.74$\pm$ 0.02 & 0.73$\pm$ 0.02 & 0.74$\pm$0.02 & 0.59 $\pm$ 0.05 & \bf{0.56} $\pm$ 0.05\\
    \bottomrule
    Total      & 0.72 $\pm$ 0.03 & 0.71$\pm$ 0.04 & 0.72 $\pm$ 0.03 & 0.58 $\pm$ 0.05 & \bf{0.54} $\pm$ 0.06\\
    \specialrule{.15em}{.05em}{.05em}
    \end{tabular}
    }
\end{table}

\section{Experiments and Applications}

\myparagraph{Implementation.} We implemented our interactive procedural vegetation framework with C++ and Vulkan. We also added a path tracer based on Nvidia OptiX 8.0. 
The 2,000 synthetic trees (tree specific 3D priors $P_{\tt 3D}(\mI_{\pi}| \mI, \pi;\sigma)$ from \secref{sec:tree_prior}) were generated for 113 hours on a laptop with NVIDIA RTX 3070. Using the 3D synthetic tree models, training the 3D prior model using Zero123~\cite{liu2023zero1to3} took 89 hours using 6 NVIDIA RTX5000 GPUs, and training the 2D prior model using LoRA~\cite{hu2021lora} took 9 hours using 4 NVIDIA RTX5000 GPU devices on the San Jose dataset. Reconstructing a 3D tree model from an RGB image takes 20 minutes using a single NVIDIA RTX 3090 GPU, and it would require $\approx$23 GPU-years on the single GPU to complete our 600,000 3D tree dataset.

\myparagraph{Experiment setup.}
We conduct experiments using the AAD dataset~\cite{beery2022auto} that contains 23 cities; among these, we use the city of San Jose to benchmark our system. We select the five most common tree genera, Cupressus, Magnolia, Pinus, Ligustrum, and Cinnamomum, from the city.

\myparagraph{Baselines.}
We compare with the Radial Bounding Volume (RBV)~\cite{Li2021ToG}, a tree-specific reconstruction using a single image with Deep Learning. We also provide comparisons with recent state-of-the-art single-image 3D object reconstruction methods based on diffusion models, including Zero123~\cite{liu2023zero1to3}, Magic123~\cite{qian2023magic123}, DreamGaussian~\cite{tang2023dreamgaussian}. 

\myparagraph{Evaluation metrics.}
We evaluate our generated 3D tree models from the San Jose subset of AAD, which contains 4,400 trees with 133 genus-level categories. We use four evaluation metrics: 
\begin{enumerate}[topsep=2pt]
\item \textbf{ICTree}~\cite{ICTree} evaluates tree plausibility, focusing on their unique traits. It assesses various geometric aspects, including branch straightness and segment ratios, using 29 parameters to analyze branching structures, as discussed in works~\cite{latentLsystem, deepTree}.
\item \textbf{LPIPS}~\cite{lpips} measures image similarity between the rendered front view of the reconstructed tree model (3D tree envelope) and the input image.
\item \textbf{CLIP-Similarity}~\cite{clip} measures the similarity between images at a semantic level. We compute the average CLIP similarity between the input image and four renderings of the reconstructed tree at orientations 0$^\mathrm{o}$, 90$^\mathrm{o}$, 180$^\mathrm{o}$, and 270$^\mathrm{o}$. Higher values indicate the renderings are more similar to the input.
\item \textbf{Chamfer distance} measures the differences in an overall tree structure using lidar scanned tree point cloud data~\cite{treepartnet} and sampled point clouds from our generated 3D tree models.
\end{enumerate}

\begin{table}[t]
    \centering
    \caption{CLIP-Similarity~\cite{clip} between four views of a tree envelope shows an improvement of 45.34\%$\pm$23.86\%.}
    \label{tab:clip_table}
    \scriptsize
    \setlength{\tabcolsep}{4pt}
        \resizebox{\linewidth}{!}{%
    \begin{tabular}{cccccc}
    \specialrule{.15em}{.05em}{.05em}
    CLIP-Sim.~\cite{clip}$\uparrow$ & DGauss.~\cite{tang2023dreamgaussian} &  Magic123~\cite{qian2023magic123} & Zero123~\cite{liu2023zero1to3} & RBV~\cite{Li2021ToG} & Ours\\
    \midrule
    Cupressus  & 0.47$\pm$0.05 & 0.49$\pm$0.03 & 0.38$\pm$0.04 & 0.54 $\pm$  0.04   & \bf{0.67} $\pm$ 0.06 \\
    Magnolia   & 0.46$\pm$0.04 & 0.46$\pm$0.03 & 0.39$\pm$0.04 & 0.54 $\pm$  0.04   & \bf{0.65} $\pm$ 0.07 \\
    Pinus      & 0.38$\pm$0.04 & 0.38$\pm$0.04 & 0.34$\pm$0.03 & 0.54 $\pm$  0.04   & \bf{0.64} $\pm$ 0.08 \\
    Ligustrum  & 0.45$\pm$0.03 & 0.45$\pm$0.9 & 0.35$\pm$0.06 & 0.54 $\pm$  0.04   & \bf{0.65} $\pm$ 0.06 \\
    Cinnamomum & 0.40$\pm$0.08 & 0.42$\pm$0.05 & 0.34$\pm$0.05 & 0.54 $\pm$  0.04   & \bf{0.65} $\pm$ 0.06 \\
    \bottomrule
    Total      & 0.43$\pm$0.06 & 0.44$\pm$0.06 & 0.36$\pm$0.05 & 0.54  $\pm$ 0.04  & \bf{0.63} $\pm$ 0.07 \\
    \specialrule{.15em}{.05em}{.05em}
    \end{tabular}
    }
\end{table}

\begin{table}[t]
    \caption{A Chamfer distance (CD) from real LiDAR scanned trees shows an improvement of 32.62\%$\pm$6.44\%.}
    \label{tab:cd_table}
    \centering
    \scriptsize
    \setlength{\tabcolsep}{4pt}
    \begin{tabular}{cccccc}
    \specialrule{.15em}{.05em}{.05em}
    CD$(10^{-2}) \downarrow$ & DGauss.~\cite{tang2023dreamgaussian} &  Magic123~\cite{qian2023magic123} & Zero123~\cite{liu2023zero1to3} & RBV~\cite{Li2021ToG} & Ours\\
    \midrule
    Tree 1  & 5.09 & 6.05 &  5.31 & 6.44 & \bf{4.62}  \\
    Tree 2  & 5.13  & 4.53 & 6.09 & 4.71  & \bf{2.81}  \\
    Tree 3  & 2.52  & 3.87 & 4.13 & 4.36  & \bf{2.31}  \\
    \bottomrule
    Total      & 4.25  & 4.82 &  5.18 & 5.17  & \bf{3.25}  \\
    \specialrule{.15em}{.05em}{.05em}
    \end{tabular}

\end{table}

\begin{figure*}[t]
    \scriptsize
    \centering
    \resizebox{0.97\linewidth}{!}{%
    \begin{tabular}{ccccccc}
        \specialrule{.15em}{.05em}{.05em}
        Genus  & Input &  RBV~\cite{Li2021ToG} & Zero123~\cite{liu2023zero1to3} & Magic123~\cite{qian2023magic123} & DGauss.~\cite{tang2023dreamgaussian} & Ours \\
        \midrule
        Cupressus & 
        \raisebox{-.5\height}{\includegraphics[width=0.85cm]{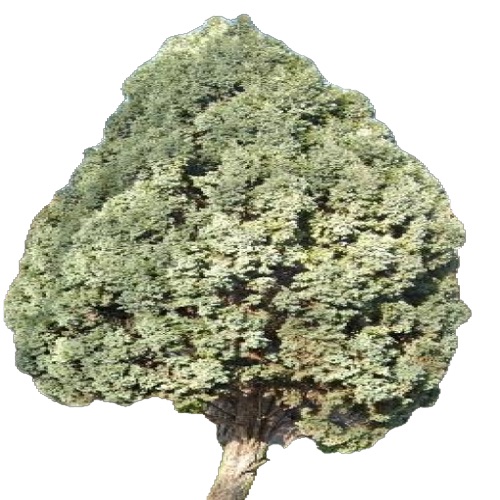}} & 
        \raisebox{-.5\height}{\includegraphics[width=0.85cm]{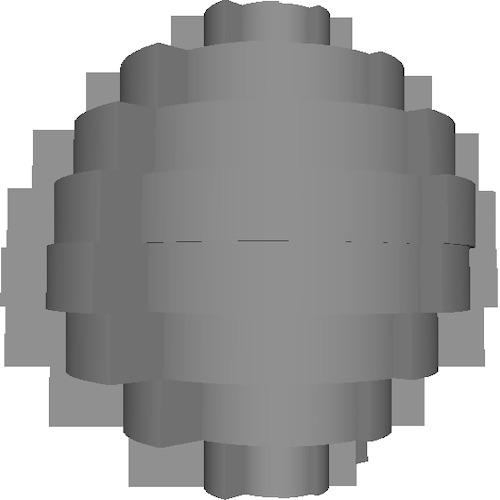}} & 
        \raisebox{-.5\height}{\includegraphics[width=0.85cm]{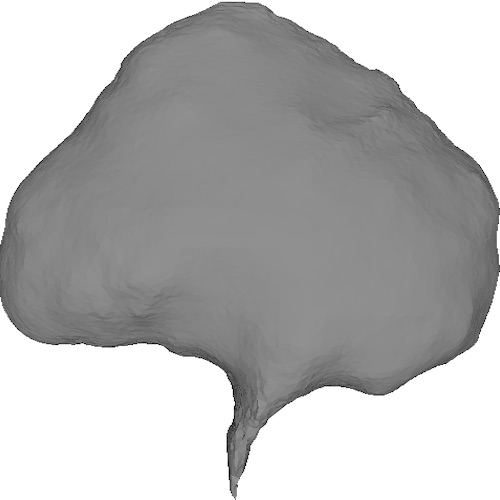}} & 
        \raisebox{-.5\height}{\includegraphics[width=0.85cm]{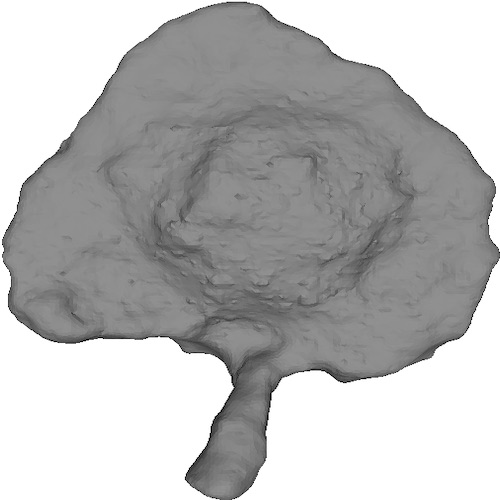}} & 
        \raisebox{-.5\height}{\includegraphics[width=0.85cm]{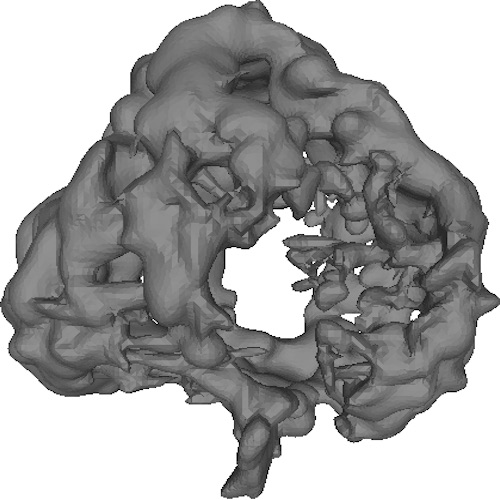}} & 
        \raisebox{-.5\height}{\includegraphics[width=0.85cm]{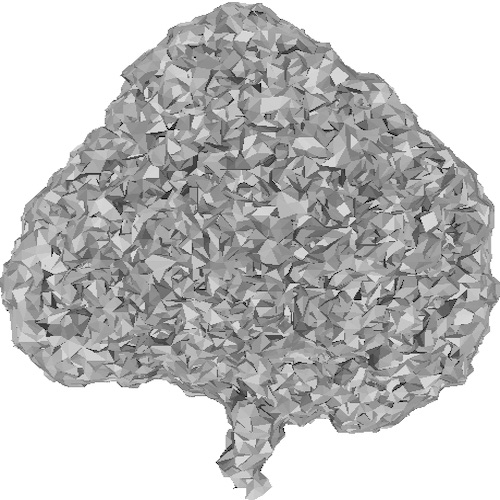}} \\
        
        Ginkgo & 
        \raisebox{-.5\height}{\includegraphics[width=0.85cm]{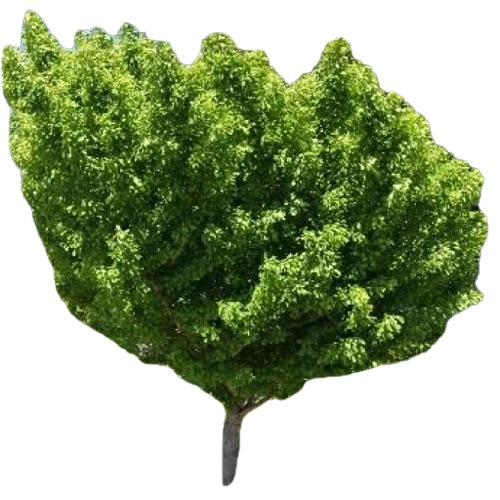}} & 
        \raisebox{-.5\height}{\includegraphics[width=0.85cm]{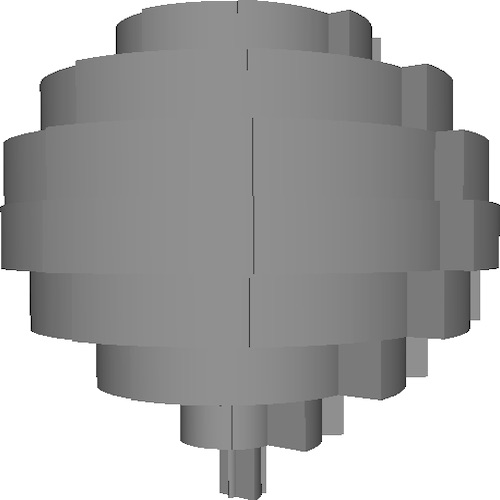}} & 
        \raisebox{-.5\height}{\includegraphics[width=0.85cm]{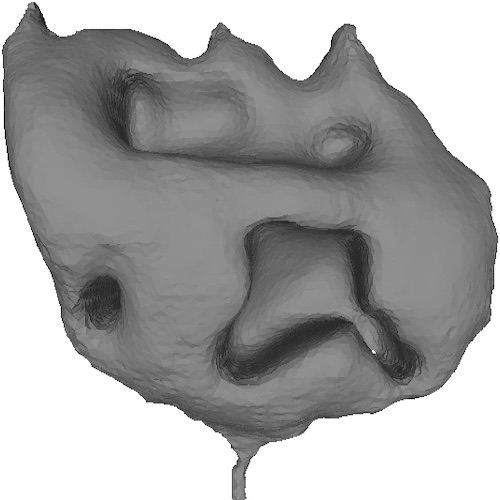}} & 
        \raisebox{-.5\height}{\includegraphics[width=0.85cm]{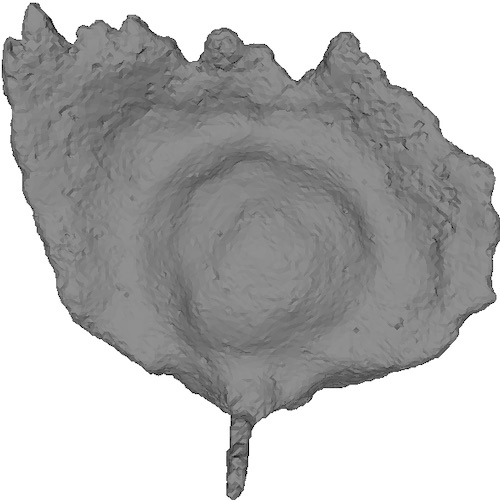}} & 
        \raisebox{-.5\height}{\includegraphics[width=0.85cm]{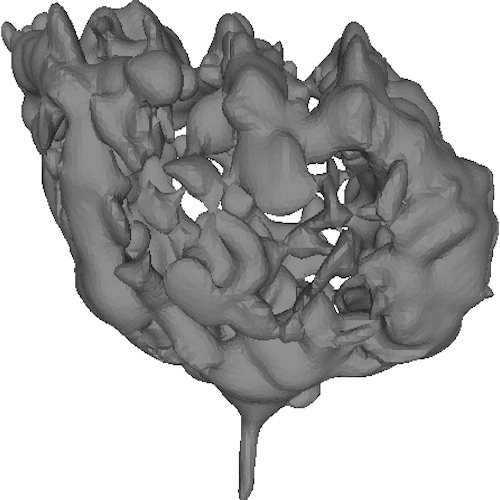}} & 
        \raisebox{-.5\height}{\includegraphics[width=0.85cm]{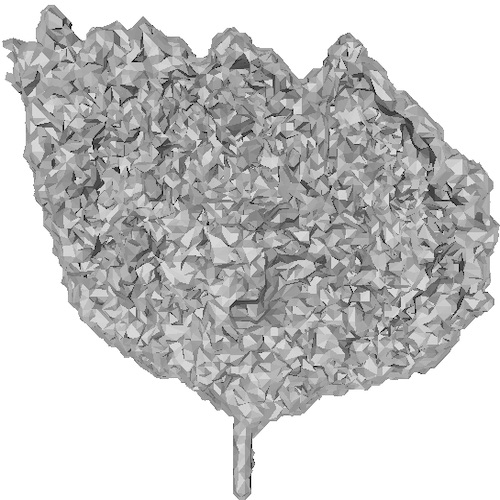}} \\
        
        Ligustrum & 
        \raisebox{-.5\height}{\includegraphics[width=0.85cm]{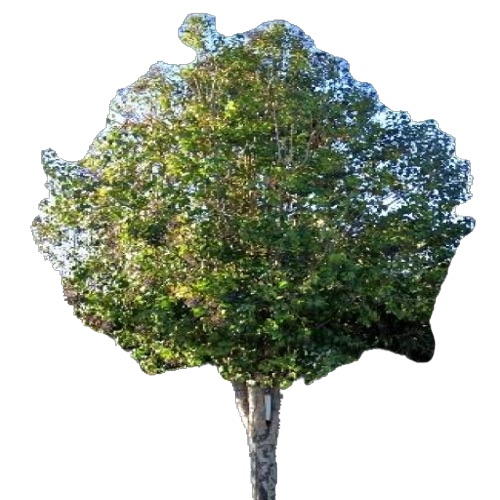}} & 
        \raisebox{-.5\height}{\includegraphics[width=0.85cm]{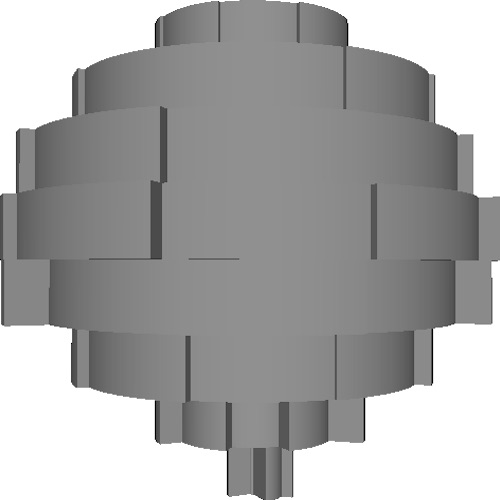}} & 
        \raisebox{-.5\height}{\includegraphics[width=0.85cm]{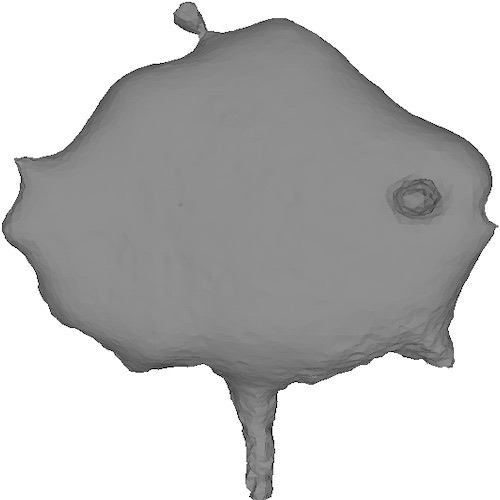}} & 
        \raisebox{-.5\height}{\includegraphics[width=0.85cm]{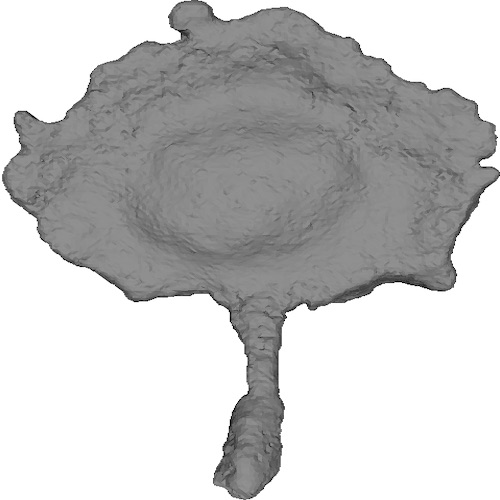}} & 
        \raisebox{-.5\height}{\includegraphics[width=0.85cm]{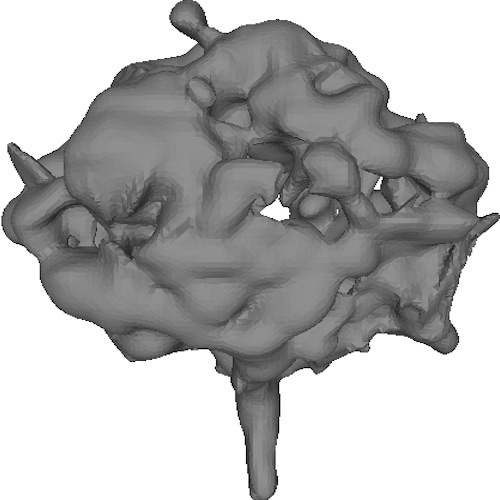}} & 
        \raisebox{-.5\height}{\includegraphics[width=0.85cm]{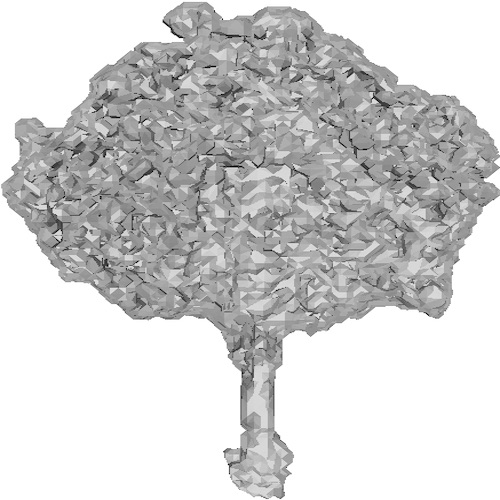}} \\
        
        Magnolia & 
        \raisebox{-.5\height}{\includegraphics[width=0.85cm]{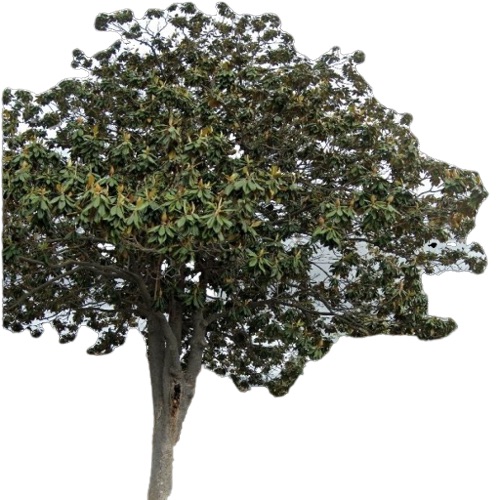}} & 
        \raisebox{-.5\height}{\includegraphics[width=0.85cm]{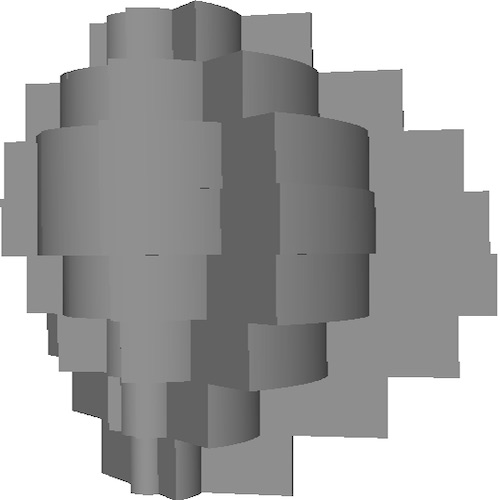}} & 
        \raisebox{-.5\height}{\includegraphics[width=0.85cm]{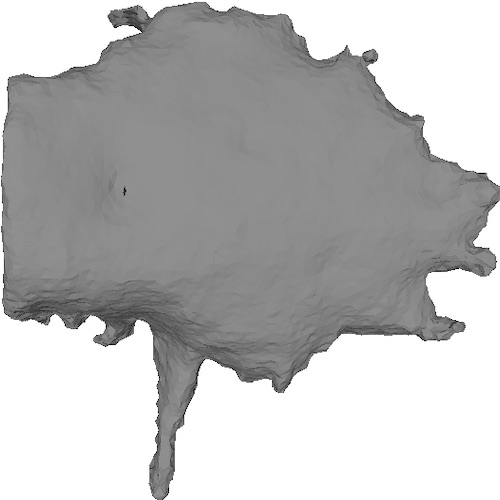}} & 
        \raisebox{-.5\height}{\includegraphics[width=0.85cm]{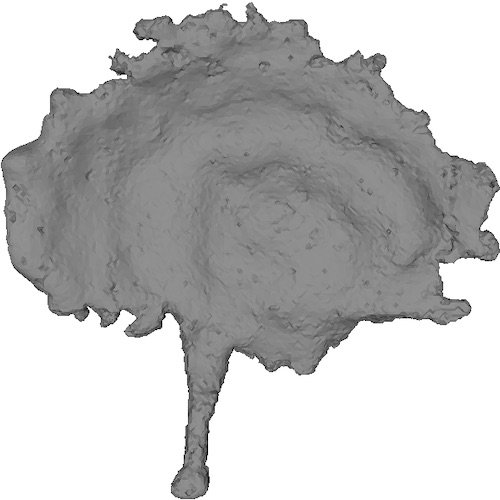}} & 
        \raisebox{-.5\height}{\includegraphics[width=0.85cm]{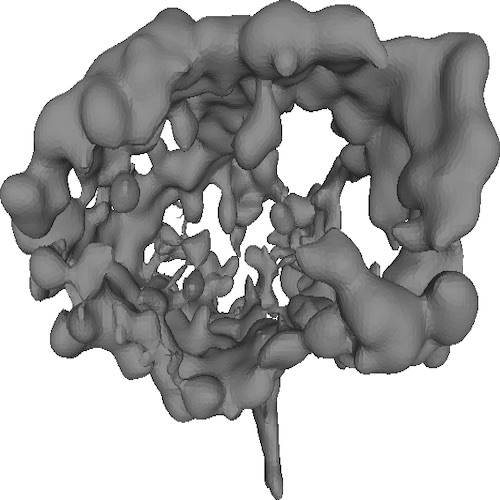}} & 
        \raisebox{-.5\height}{\includegraphics[width=0.85cm]{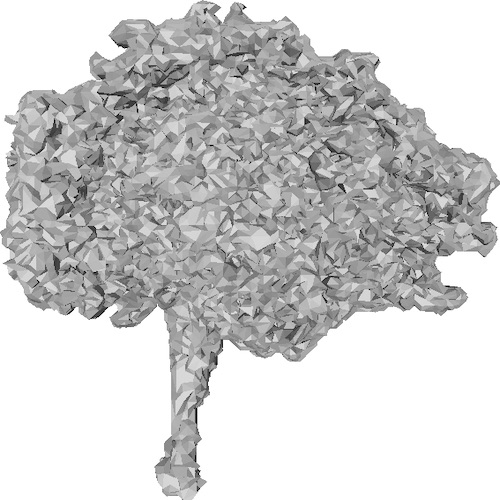}} \\
        
        Tristaniopsis & 
        \raisebox{-.5\height}{\includegraphics[width=0.85cm]{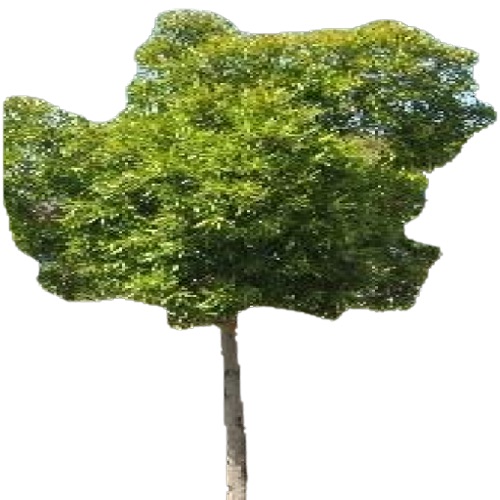}} & 
        \raisebox{-.5\height}{\includegraphics[width=0.85cm]{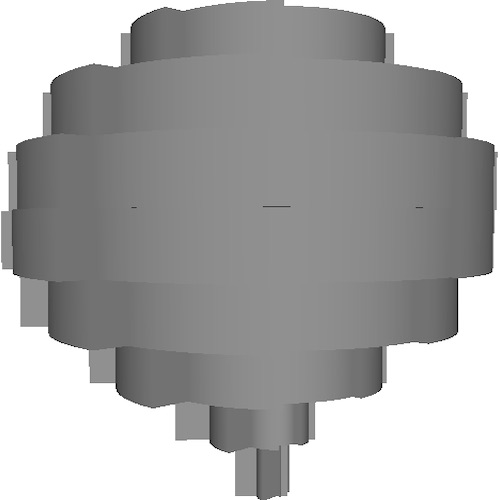}} & 
        \raisebox{-.5\height}{\includegraphics[width=0.85cm]{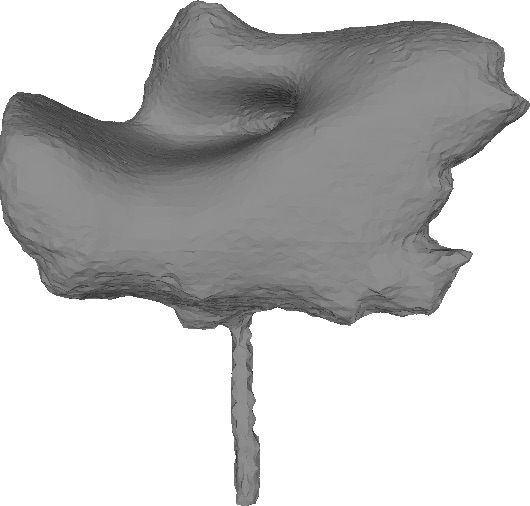}} & 
        \raisebox{-.5\height}{\includegraphics[width=0.85cm]{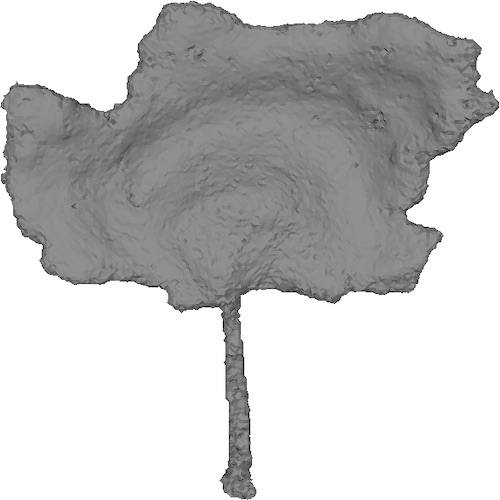}} & 
        \raisebox{-.5\height}{\includegraphics[width=0.85cm]{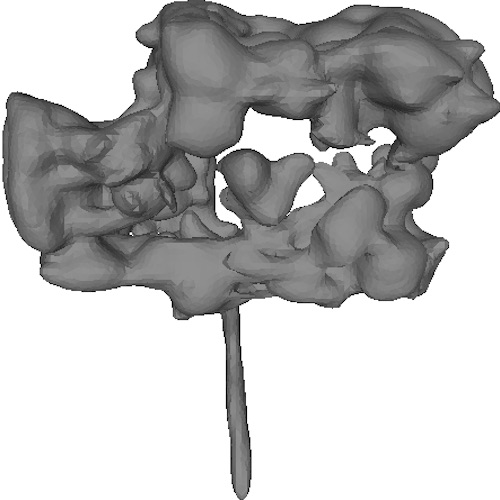}} & 
        \raisebox{-.5\height}{\includegraphics[width=0.85cm]{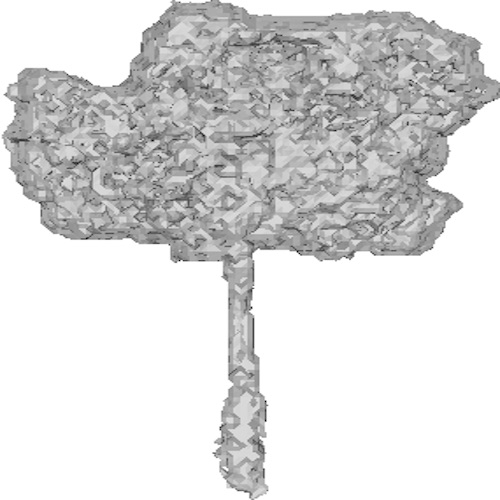}} \\
        \specialrule{.15em}{.05em}{.05em}
    \end{tabular}
    }
    \caption{Single tree image reconstruction front-view results.}
    \label{fig:baseline_front_view}
\end{figure*}

\begin{figure*}[t]
    \scriptsize
    \centering
    \resizebox{0.97\linewidth}{!}{%
    \begin{tabular}{ccccccc}
        \specialrule{.15em}{.05em}{.05em}
        Genus  & Input &  RBV~\cite{Li2021ToG} & Zero123~\cite{liu2023zero1to3} & Magic123~\cite{qian2023magic123} & DGauss.~\cite{tang2023dreamgaussian} & Ours \\
        \midrule
        Cupressus & 
        \raisebox{-.5\height}{\includegraphics[width=0.85cm]{data/experiment/reconstructions/ours/cupressus/cupressus_input_bg.jpeg}} & 
        \raisebox{-.5\height}{\includegraphics[width=0.85cm]{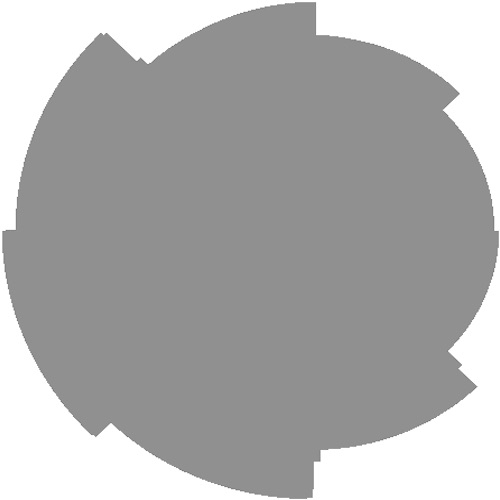}} & 
        \raisebox{-.5\height}{\includegraphics[width=0.85cm]{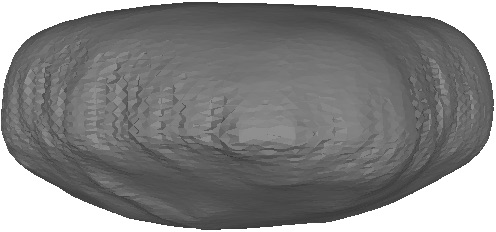}} & 
        \raisebox{-.5\height}{\includegraphics[width=0.85cm]{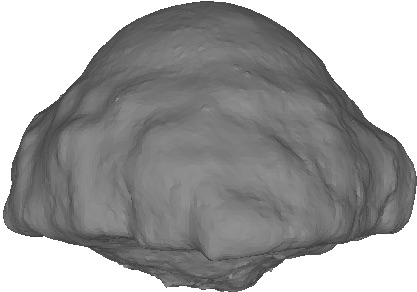}} & 
        \raisebox{-.5\height}{\includegraphics[width=0.85cm]{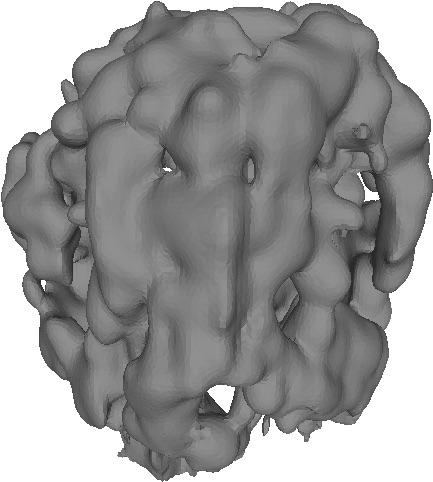}} & 
        \raisebox{-.5\height}{\includegraphics[width=0.85cm]{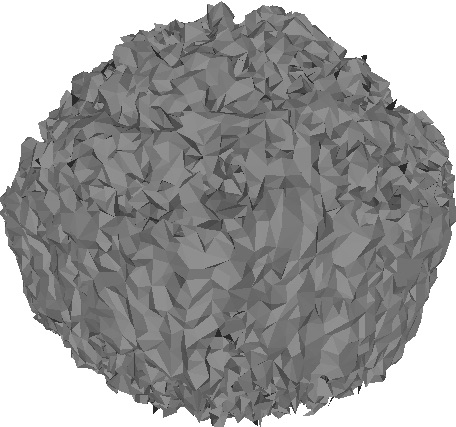}} \\
        
        Ginkgo & 
        \raisebox{-.5\height}{\includegraphics[width=0.85cm]{data/experiment/reconstructions/ours/ginkgo/ginkgo_input_bg.jpeg}} & 
        \raisebox{-.5\height}{\includegraphics[width=0.85cm]{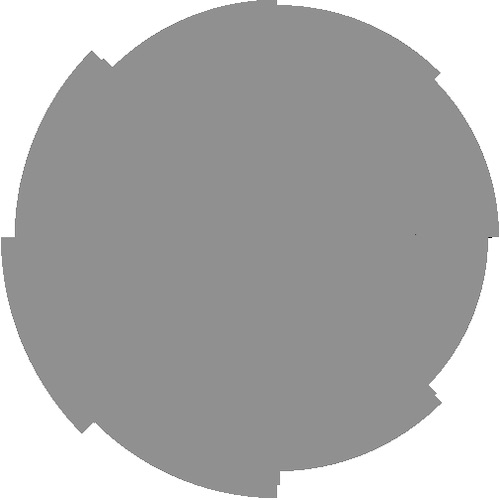}} & 
        \raisebox{-.5\height}{\includegraphics[width=0.85cm]{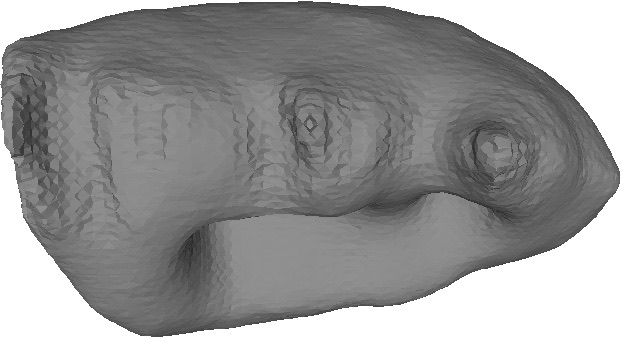}} & 
        \raisebox{-.5\height}{\includegraphics[width=0.85cm]{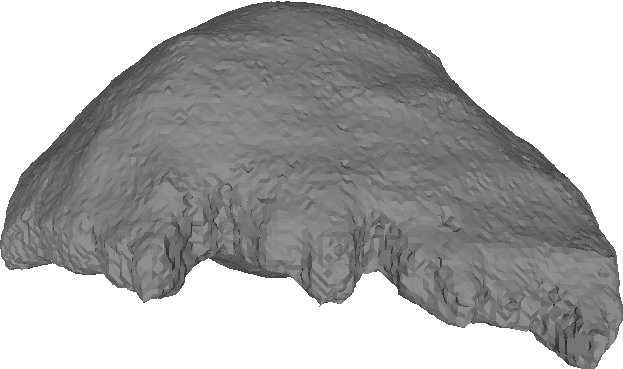}} & 
        \raisebox{-.5\height}{\includegraphics[width=0.85cm]{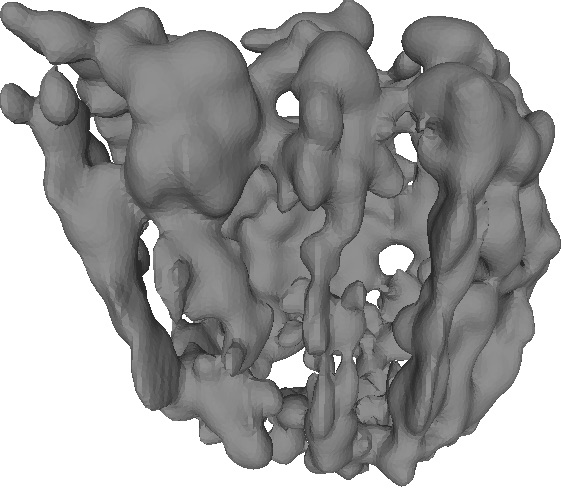}} & 
        \raisebox{-.5\height}{\includegraphics[width=0.85cm]{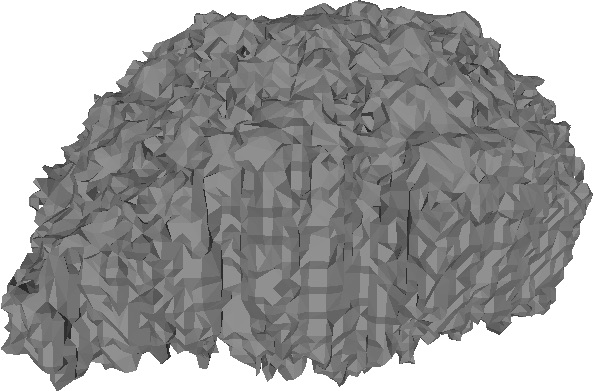}} \\
        
        Ligustrum & 
        \raisebox{-.5\height}{\includegraphics[width=0.85cm]{data/experiment/reconstructions/ours/ligustrum/ligustrum_input_bg.jpeg}} & 
        \raisebox{-.5\height}{\includegraphics[width=0.85cm]{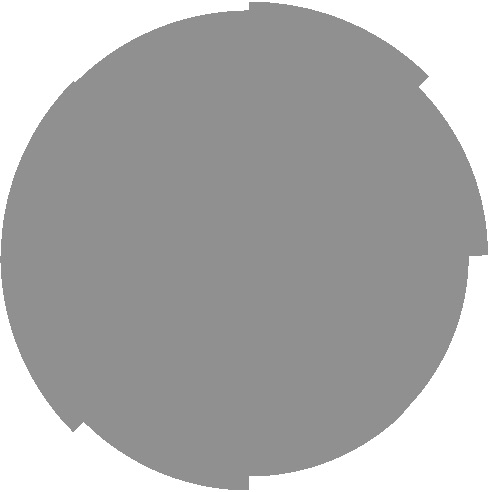}} & 
        \raisebox{-.5\height}{\includegraphics[width=0.85cm]{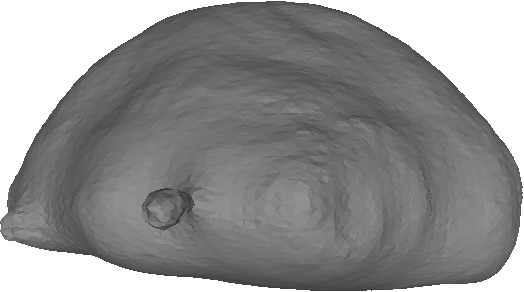}} & 
        \raisebox{-.5\height}{\includegraphics[width=0.85cm]{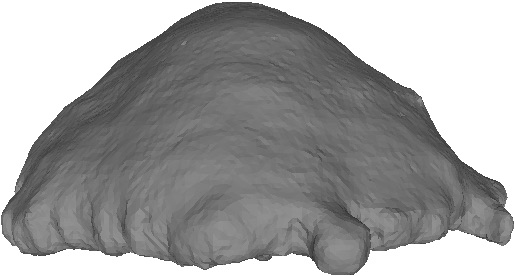}} & 
        \raisebox{-.5\height}{\includegraphics[width=0.85cm]{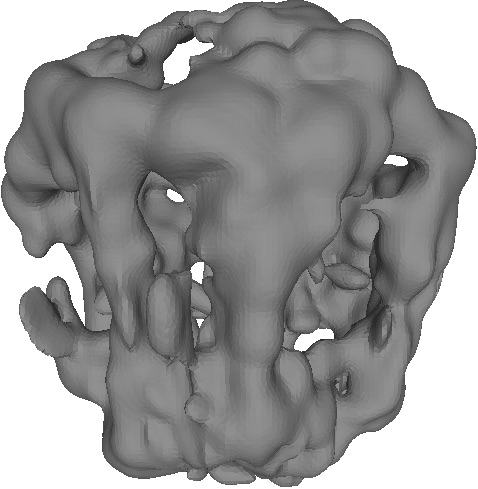}} & 
        \raisebox{-.5\height}{\includegraphics[width=0.85cm]{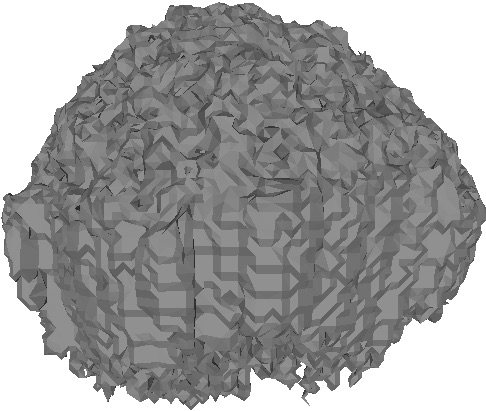}} \\
        
        Magnolia & 
        \raisebox{-.5\height}{\includegraphics[width=0.85cm]{data/experiment/reconstructions/ours/magnolia/magnolia_input_bg.jpeg}} & 
        \raisebox{-.5\height}{\includegraphics[width=0.85cm]{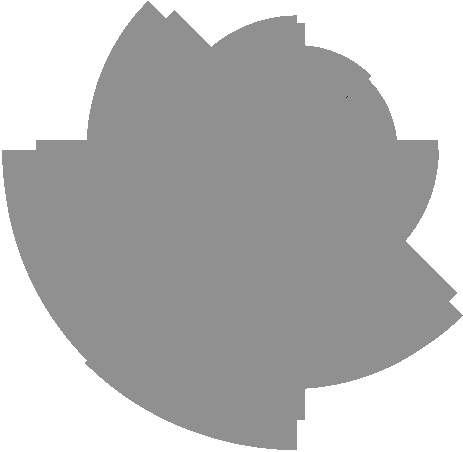}} & 
        \raisebox{-.5\height}{\includegraphics[width=0.85cm]{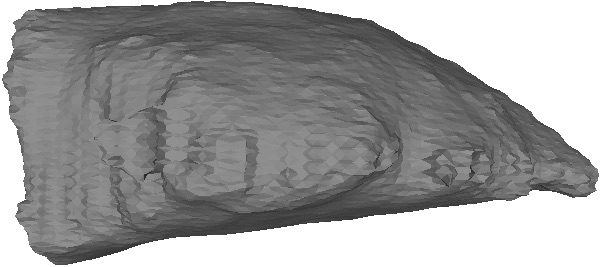}} & 
        \raisebox{-.5\height}{\includegraphics[width=0.85cm]{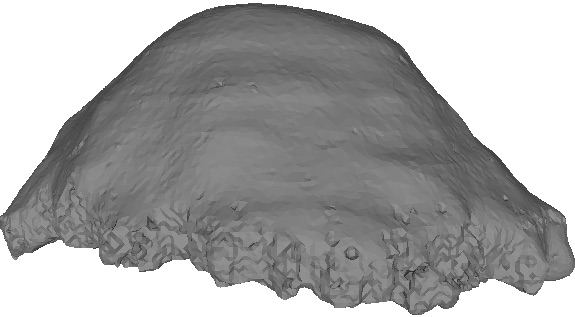}} & 
        \raisebox{-.5\height}{\includegraphics[width=0.85cm]{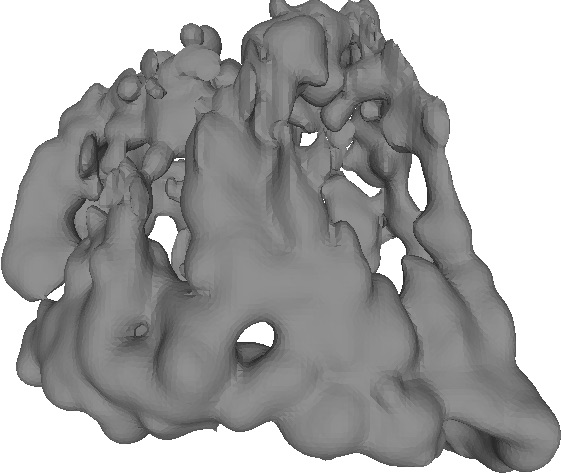}} & 
        \raisebox{-.5\height}{\includegraphics[width=0.85cm]{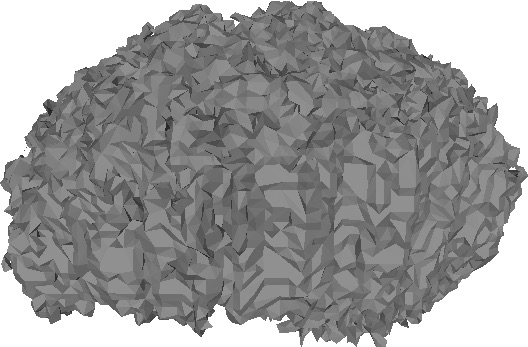}} \\
        
        Tristaniopsis & 
        \raisebox{-.5\height}{\includegraphics[width=0.85cm]{data/experiment/reconstructions/ours/tristaniopsis/tristaniopsis_input_bg.jpeg}} & 
        \raisebox{-.5\height}{\includegraphics[width=0.85cm]{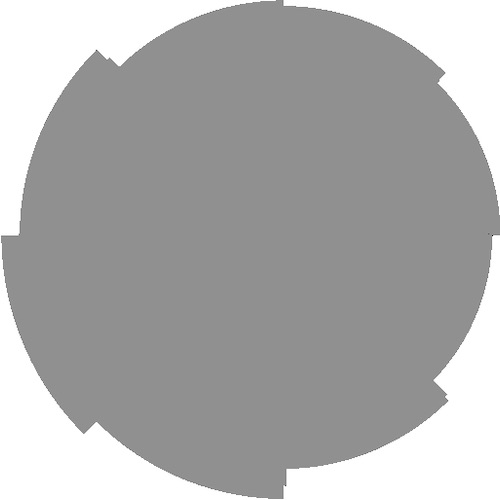}} & 
        \raisebox{-.5\height}{\includegraphics[width=0.85cm]{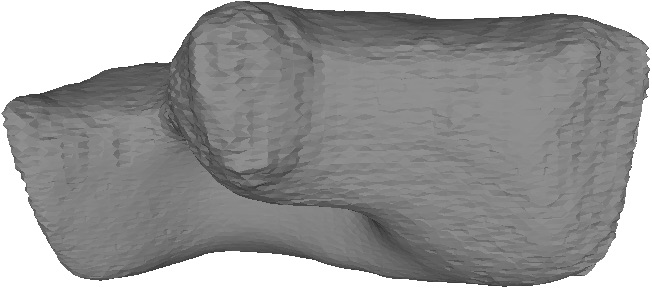}} & 
        \raisebox{-.5\height}{\includegraphics[width=0.85cm]{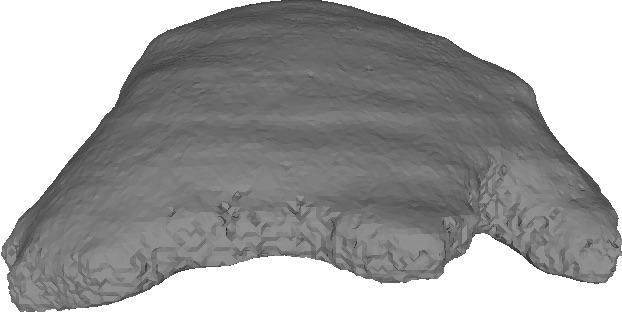}} & 
        \raisebox{-.5\height}{\includegraphics[width=0.85cm]{data/experiment/reconstructions/dreamgaussian/cupressus/top01.jpeg}} & 
        \raisebox{-.5\height}{\includegraphics[width=0.85cm]{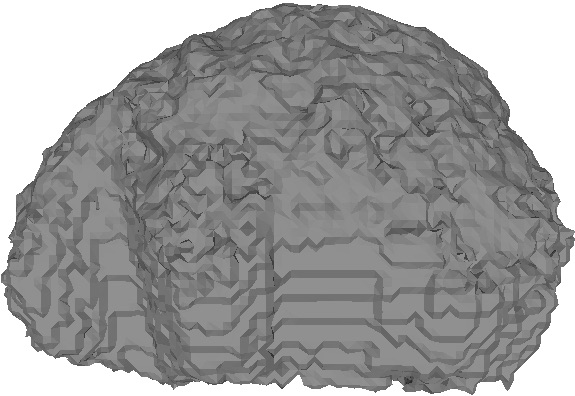}} \\
        \specialrule{.15em}{.05em}{.05em}     
    \end{tabular}
    }
    \caption{Single tree image reconstruction top-view results.}
    \label{fig:baseline_top_view}
\end{figure*}

\subsection{Quantitative Results}
\label{sec:quant_results}
\tabref{tab:ictree_table} reports the ICTree~\cite{ICTree} score to compare the perceptual realism of tree branching structures generated using \name. We use 133 tree genera from the San Jose dataset, and on average, \name\ scores 0.71 with a standard deviation of 0.05, which is an improvement in the average of 20.21\%$\pm$8.89\% compared to the recent works~\cite{tang2023dreamgaussian, qian2023magic123, liu2023zero1to3, Li2021ToG}. \name\ achieves unparalleled realism in generating tree structures by confirming 29 geometric branching attributes.

In~\tabref{tab:lpips_table}, we use the pre-trained VGG~\cite{vgg} network to calculate a perceptual similarity between our generated and RBV tree front-views. We show the LPIPS score of five common tree genera from the San Jose dataset, including all the tree genera in~\tabref{tab:lpips_table}. Ours shows 20.21\%$\pm$8.89\% greater perceptual similarity to a conditioned image than the generated trees from the previous works~\cite{tang2023dreamgaussian, qian2023magic123, liu2023zero1to3, Li2021ToG}.

\tabref{tab:clip_table} shows the CLIP-Similarity between RBV~\cite{Li2021ToG} and \name. We used four evenly distributed views around the object's 360-degree to validate the consistency of the 3D-generated tree model. Using the five common tree genera from the San Jose dataset, \name\ outperforms other works~\cite{tang2023dreamgaussian, qian2023magic123, liu2023zero1to3, Li2021ToG} by an average of 45.34\%$\pm$23.86\%.

\tabref{tab:cd_table} compares the Chamfer Distance between LiDAR-scanned tree point clouds and our results using captured images~\cite{treepartnet}. We show that our generated tree models are reconstructed in the average of 32.62\%$\pm$6.44\% closer to the real trees than other works~\cite{tang2023dreamgaussian, qian2023magic123, liu2023zero1to3, Li2021ToG}.

Based on the four reconstruction metrics, \name\ shows it becomes a new state-of-the-art method to reconstruct a 3D tree model from a single unposed RGB image. Thus, our dataset contains 3D tree models that are perpetually, visually, and geometrically realistic.

\subsection{Qualitative Results}
We show the existing methods generate unsatisfying tree structures from a single image due to a lack of tree geometry information and visualize its front~(\figref{fig:baseline_front_view}) and top-views~(\figref{fig:baseline_top_view}).
RBV~\cite{Li2021ToG} capture the approximated contour estimation, Zero123~\cite{liu2023zero1to3} also captures contour information but includes holes in the Ginko and Ligustrum genus, and the front-view is flat. Also, it generates non-tree-like shapes near the top area in the Tristaniopsis tree genus. Magic123~\cite{qian2023magic123} fails to estimate tree depth, concentrating on the center and creating a gap between the inner and outer sides. Moreover, it does not generate overall balanced tree volumes and tends to look overly simplistic. DreamGaussian~\cite{tang2023dreamgaussian} captures the overall tree contour but it creates numerous holes around the center of 3D tree models with bubble-like artifacts visible from the top view. \name\ captures the tree contours without any holes and generates tree-like balanced volumetric trees from the front and the top view.

\subsection{Ablation Study}
\begin{wrapfigure}[9]{l}{0.2\textwidth} 
\centering
\includegraphics[width=\linewidth]{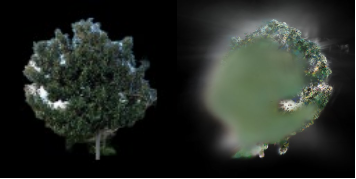}
\caption{Novel view failed using 3DGS.}
\label{fig:rbt_3dgs}
\end{wrapfigure}

We conduct ablation studies using 3D Gaussian Splatting~\cite{kerbl20233d} on a tree and different $\alpha$/$\beta$ ratios. Synthesizing novel views for trees, as seen in~\cref{fig:rbt_3dgs},
\vspace{-0.45cm}

\noindent has proven challenging, hindering the extraction of high-quality 3D boundaries. Successful synthesis relies on pre-trained data, as shown by our use of 24,000 synthetic 3D tree models. This highlights the need for large-scale real image-based 3D tree reconstruction datasets in forestry to utilize deep learning efficiently.
\begin{table}[hbt]
    \caption{A reconstruction quality comparison between different $\alpha$/$\beta$ ratios.}
    \label{tab:rbt_ablation}
    \centering
    \setlength{\tabcolsep}{4pt}
    \begin{tabular}{cccc}
    \specialrule{.15em}{.05em}{.05em}
    $\alpha/\beta$& CLIP~\cite{clip}$\uparrow$ & LPIPS~\cite{lpips}$\downarrow$ & ICTree~\cite{ICTree}$\uparrow$\\
    \midrule
     0.01 & 0.53 $\pm$ 0.06 & 0.58 $\pm$ 0.07 & 0.52 $\pm$ 0.07\\   
     0.1 & 0.60 $\pm$ 0.07 & 0.59$\pm$ 0.06 & 0.55 $\pm$ 0.06\\   
     1 & 0.59$\pm$ 0.07 & 0.58$\pm$ 0.07 & 0.50 $\pm$ 0.05\\ 
     10 & 0.57$\pm$ 0.07 & 0.59$\pm$ 0.06 & 0.47$\pm$ 0.03\\
    \specialrule{.15em}{.05em}{.05em}
    \end{tabular}
\end{table}

We evaluate the tree reconstruction metrics, CLIP-Similarity~\cite{clip}, LPIPS~\cite{lpips} and ICTree~\cite{ICTree} using four different $\alpha$/$\beta$ ratios. Empirically, from~\cref{tab:rbt_ablation}, we found that the ratio around 0.1 achieves a good performance.

\section{Conclusion}\label{sec:concl}
We introduce \name, the first large dataset of 3D reconstructed tree models enabled by a diffusion-based approach to construct 3D tree models from single images. We use images from the AAD~\cite{beery2022auto} and synthetic trees to train diffusion models that reconstruct 3D tree volumes that are then completed by a developmental model to provide functional-structural tree model for individual real trees. Our generated dataset has the potential to enable diverse and impactful applications in 3D tree phenotyping at a previously unprecedented scale. 

\begin{wrapfigure}[13]{r}{0.3\textwidth} 
\centering
\includegraphics[height=2.0cm]{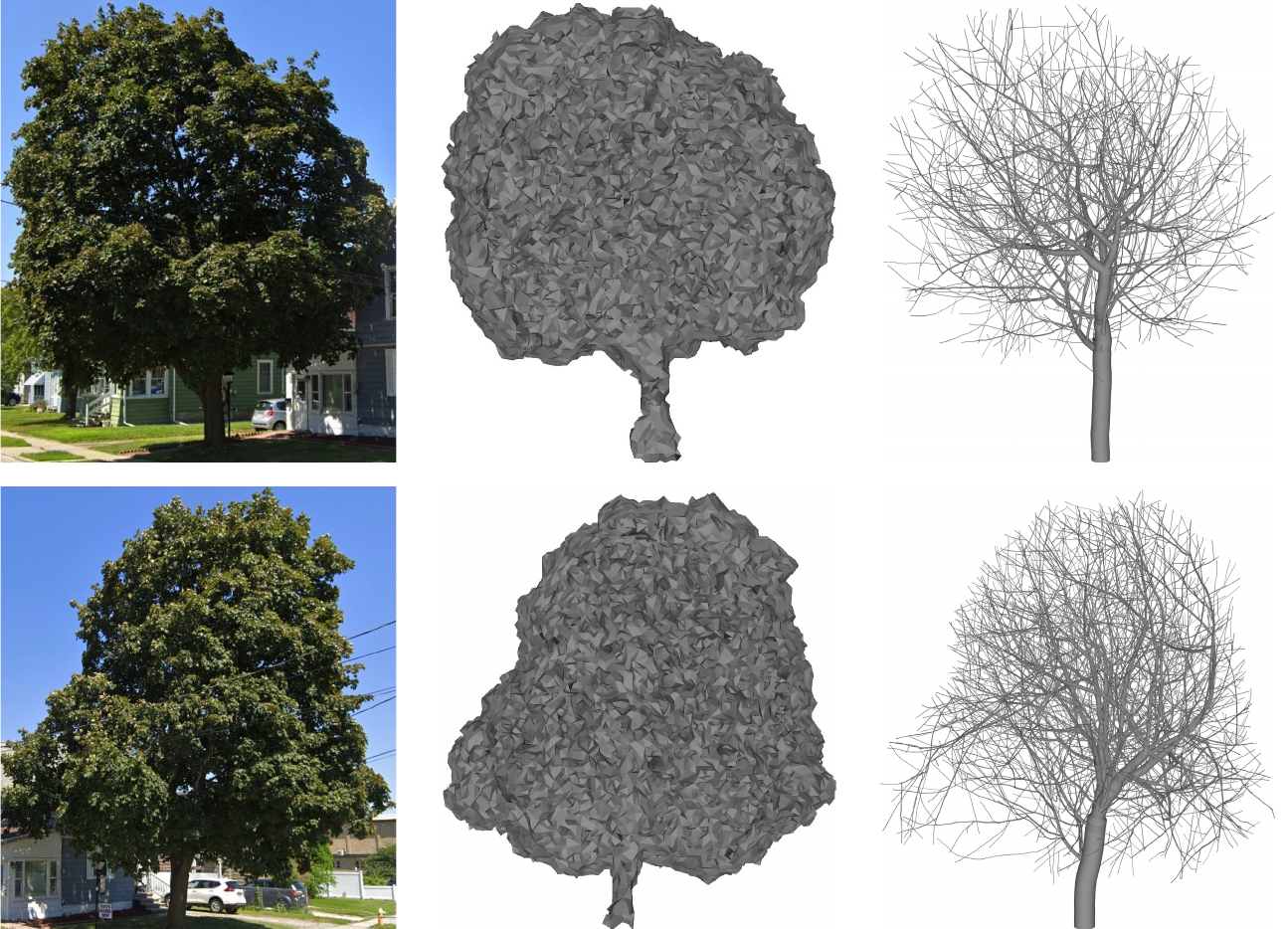}
\caption{Single-image reconstruction fails with asymmetric trees, producing different 3D models from varying views.}
\label{fig:our_failure}
\end{wrapfigure}

This is the first large dataset of its kind, but it comes with a few \textbf{limitations}.
First, the 3D model shapes are constrained by the capabilities of the simulator. Second, single-image reconstruction, limited by a single viewpoint and missing information, is best suited for symmetric trees.
An example in~\figref{fig:our_failure} epitomizes this case.
The same asymmetric tree from two different angles yields significantly varied 3D models. High branch density can also hinder precise reconstruction due to the complexity of tree segmentation.
Finally, predicting a branch structure obscured by leaves with 100\% fidelity to nature is an ongoing challenge due to its complex geometry.

There are many possible avenues for future work.
First, a comprehensive 3D tree database could enable the study of light and environmental effects on reconstruction. Additionally, incorporating real tree models with algorithms for partially occluded parts during training could improve detection significantly, as single-image segmentations struggle to disentangle nearby tree crowns.

\section*{Acknowledgements}
This work is based upon efforts supported by the EFFICACI grant under USDA NRCS \#NR233A750004G044. The views and conclusions contained herein are those of the authors and should not be interpreted as representing the official policies, either expressed or implied, of the U.S. Government or NRCS. The U.S. Government is authorized to reproduce and distribute reprints for governmental purposes notwithstanding any copyright annotation therein. This project was also sponsored by USDA NIFA, Award \#2023-68012-38992 grant ``Promoting Economic Resilience and Sustainability of the Eastern U.S. Forests'' and by NIFA-AFRI award \#2024-67021-42879, “Improving Forest Management Through Automated Measurement of Tree Geometry”.

\clearpage

\bibliographystyle{splncs04nat}
\bibliography{ref,sara,shape}

\end{document}